\makeatletter
\providecommand\Notice@String[1]{#1}
\makeatother

\documentclass{article}

\usepackage{microtype}
\usepackage{graphicx}
\usepackage{subcaption}
\usepackage{booktabs} 

\usepackage{hyperref}

\usepackage[accepted]{mlsys2025}

\usepackage{enumitem}
\usepackage{booktabs}
\usepackage{inconsolata}
\usepackage{subcaption}
\usepackage{amsmath}
\usepackage{listings}

\definecolor{codebg}{RGB}{248,248,248}
\definecolor{coderule}{RGB}{215,215,215}
\lstdefinestyle{codebox}{
  basicstyle=\ttfamily\scriptsize,
  backgroundcolor=\color{codebg},
  frame=single,
  rulecolor=\color{coderule},
  framerule=0.4pt,
  columns=fullflexible,
  keepspaces=true,
  breaklines=true,
  breakatwhitespace=false,
  showstringspaces=false,
  tabsize=4,
  aboveskip=0.3em,
  belowskip=0.0em,
  xleftmargin=0.35em,
  xrightmargin=0.15em,
}

\usepackage{quoting}

\mlsystitlerunning{Speculative Decoding: Performance or Illusion?}

\begin{document}

\twocolumn[
\mlsystitle{Speculative Decoding: Performance or Illusion?}



\mlsyssetsymbol{equal}{*}

\begin{mlsysauthorlist}
\mlsysauthor{Xiaoxuan Liu}{equal,to}
\mlsysauthor{Jiaxiang Yu}{equal,to}
\mlsysauthor{Jongseok Park}{to}
\mlsysauthor{Ion Stoica}{to}
\mlsysauthor{Alvin Cheung}{to}
\end{mlsysauthorlist}

\mlsysaffiliation{to}{UC Berkeley}

\mlsyscorrespondingauthor{Xiaoxuan Liu}{xiaoxuan\_liu@berkeley.edu}
\mlsyscorrespondingauthor{Jiaxiang Yu}{jxyu@berkeley.edu}

\mlsyskeywords{Machine Learning, MLSys}

\vskip 0.3in

\begin{abstract}
Speculative decoding (SD) has become a popular technique to accelerate Large Language Model (LLM) inference, yet its real-world effectiveness remains unclear as prior evaluations rely on research prototypes and unrealistically small batch sizes. We present, to our knowledge, the first systematic study of SD on a production-grade and widely deployed inference engine (vLLM), covering multiple SD variants ($n$-gram, EAGLE/EAGLE-3, Draft-Model, Multi-Token Prediction) across diverse workloads, model scales, and batch sizes. We analyze key factors governing SD performance, and 
quantify a theoretical upper bound on SD speedup. Our results show that verification by the target model dominates the execution, while acceptance length varies markedly across output token positions, requests, and datasets. Comparing measured performance with theoretical bounds reveals substantial gaps between observed and theoretical upper bounds, and we leverage this observation to highlight new research opportunities that our study opens up in improving SD. Our code for profiling and simulator is available at \href{https://github.com/orgs/SpecDecode-Bench/repositories}{https://github.com/orgs/SpecDecode-Bench/repositories}.
\end{abstract}
]



\printAffiliationsAndNotice{\mlsysEqualContribution} 

\section{Introduction}
Speculative decoding (SD) has emerged as a highly effective approach to speed up inference in large language models (LLMs).
It has been widely used in real workloads, ranging from chat completion~\cite{chatgpt}, question answering~\cite{cobbe2021gsm8k, rein2023gpqagraduatelevelgoogleproofqa} to coding tasks~\cite{jimenezswe}. 

Despite the substantial progress in SD, there has not been a systematic study on its effectiveness.
Prior studies only used prototype implementations rather than production-level systems, which greatly undermines the validity of their conclusions. 
Worse yet, these evaluations are often conducted with a batch size of one~\cite{xia-etal-2024-unlocking, cai2024medusa, leviathan2023fast, li2024eagle, li2024eagle2fasterinferencelanguage, li2025eagle3}—an unrealistic configuration that fails to reflect real-world deployment.
Second, since the introduction of SD, numerous variants have emerged, including draft-model-based~\cite{miao2023specinfer, liu2023online, zhou2023distillspec, cascade-inference, chen2024sequoia}, $n$-gram-based~\cite{saxena2023prompt, somasundaram2024pld+}, and tree-based~\cite{cai2024medusa, li2024eagle, lin2024bita, fu2024break, li2024eagle2fasterinferencelanguage, li2025eagle3} approaches. Yet, we are unaware of any systematic analysis comparing these variants and understanding their respective use cases. This gap poses practical challenges: when deploying SD in production, practitioners are often left without clear guidance on which variant to use for different model architectures or workload conditions.

We aim to study SD in actual deployment settings using real-world workloads. Towards that end, we systematically benchmark SD in vLLM~\cite{kwon2023efficient}, a production-ready inference system, and evaluate multiple SD variants across diverse datasets. 
Although this setup may appear straightforward, several subtle yet important caveats arise in practice. 
First, while SD is theoretically guaranteed to preserve the same token distribution as standard decoding, we observe that the generated outputs are not always identical. This discrepancy stems from the inherent nondeterminism~\cite{he2025nondeterminism} in LLM inference—caused by factors such as kernel parallelism and floating-point variation. 
Furthermore, we study the SD performance on two increasingly prominent setups: reasoning workloads, which involve longer and more structured generations, and multi-token prediction (MTP)~\cite{deepseekai2025deepseekv3technicalreport, 5team2025glm45agenticreasoningcoding}, a recently emerging acceleration technique that complements SD.
Across all workloads, every SD variant outperforms the no-SD baseline. The speedup decreases as the batch size increases, consistent with the reduced opportunities for SD on computation-bound scenarios~\cite{liu2024dsd}. Notably, reasoning workloads that require long chains of thought—and therefore generate longer outputs—exhibit speedups comparable to those observed on standard language-modeling tasks.

Next, we conduct a detailed analysis to understand the performance of speculative decoding (SD) across different variants. The overall speedup of SD is governed by two key factors: the execution efficiency of its constituent stages and the token acceptance rate.

We first examine the execution efficiency of the different SD stages.
In general, SD consists of three stages---proposing tokens, verifying tokens, and system overhead such as scheduling and rejection sampling. 
By examining the time breakdown of the three stages, we find that the proposing stage accounts for only a small portion of the total execution time, and the execution of the large model during verification remains the dominant cost. Thus, when the acceptance rate is low, repeatedly running the large model on rejected tokens incurs substantial runtime cost. As shown in prior work~\cite{liu2024dsd}, this extra cost becomes prohibitive under high system load, and can even cause SD to be slower than standard decoding.

We next conduct a detailed analysis of acceptance behavior in SD, examining when proposed tokens are accepted or rejected.
We observe that acceptance patterns exhibit three levels of variability: within a request, across different requests, and across different datasets. 
Furthermore, different SD methods demonstrate entirely different behavior. For example, learned SD methods such as EAGLE maintain stable and consistently high acceptance across diverse workloads, whereas training-free method like $n$-gram produces more variable but occasionally much longer accepted spans in tasks like code-editing, where frequent local repetitions allow it to reuse previously seen patterns.

Based on the above observations, we ask a fundamental question: \emph{how far are current SD approaches from the optimal speedup, and what is the theoretical upper bound of speculative decoding?}

First, motivated by the observation that a large fraction of proposed tokens are ultimately rejected, we identify a promising direction for improving SD performance: \textbf{verifying only those tokens that are likely to be accepted}. In the ideal case, a proposal method that perfectly predicts the large model’s outputs would eliminate the need to invoke the large model altogether. To quantify the gap between observed and theoretical speedups, we develop a simulator grounded in real-world benchmarking data. This simulator evaluates performance under an idealized setting in which all proposed tokens are accepted, thereby minimizing verification cost and achieving the optimal speedup.

Second, we observe that different SD methods achieve higher acceptance at different token positions. This complementarity suggests that adaptively combining multiple SD methods can unlock additional speedup beyond what any single method can achieve in isolation. Our results show that such adaptive combinations can improve end-to-end speedup to 4.9$\times$ relative to standard (no-SD) decoding.

In summary, the paper makes the following contributions:
\begin{itemize}[leftmargin=*, itemsep=2pt]
    \item \textbf{Production-grade evaluation of speculative decoding.} 
    We present the first systematic study of SD within a widely adopted, highly optimized inference engine (vLLM), thereby bridging the gap between research prototypes and real-world deployments. We evaluate mainstream SD variants across various real-world workloads.

    \item \textbf{Decomposition of speedup and performance bottlenecks.} 
    We analyze key factors that affect SD performance and find that: (1) verification cost remains dominant, and (2) acceptance behavior exhibits variability across positions, requests, and datasets. To the best of our knowledge, this is the first work to thoroughly examine both the time/memory breakdown and acceptance dynamics of SD across diverse workloads.

    \item \textbf{Theoretical upper bound of SD speedup.} 
    Based on these observations, we analyze the gap between measured and maximum achievable speedup. By leveraging position-specific acceptance statistics without modifying the proposing method, we compute the theoretical upper bound of SD performance, revealing an orthogonal and promising direction for future optimization.
\end{itemize}

\section{Background and Motivation}
\textbf{Speculative Decoding.}
Large language models (LLMs) generate text auto-regressively: given a prompt $(x_1, \ldots, x_n)$, they produce an output sequence $(x_{n+1}, \ldots, x_{n+S})$, 
one token at a time. At each step, the model computes a probability distribution over the vocabulary and samples the next token. 
However, while LLMs \emph{generate} tokens sequentially, they can also \emph{evaluate} the probabilities of a \emph{given} token sequence in parallel. That is, for candidate tokens $x_{n+1}, \ldots, x_{n+S}$, an LLM can compute $P(x_{n+1} \mid x_1, \ldots, x_n), \ldots, P(x_{n+S} \mid x_1, \ldots, x_{n+S-1})$ in a single forward pass.
Speculative decoding~\cite{leviathan2023fast, chen2023accelerating} utilizes this property of LLMs as an evaluator on a string of candidate tokens in parallel, enabling higher hardware utilization and potentially generating multiple tokens.
For example, a smaller model may generate $k$ candidate tokens, and our target LLM model then predicts $k+1$ probability distributions for each token position in parallel. We then accept the first $m$ tokens from the $k$ candidate tokens that satisfy the sampling method. 

\textbf{SD Variations.}
When SD was first introduced, the standard approach was to use a smaller draft model with the same vocabulary as the target model, since rejection sampling must operate on a probability distribution in the same token embedding space.
Subsequent work has proposed various ways to construct the draft model, such as quantization~\cite{lin2023awq, frantar2023gptq, kim2024squeezellm} or distillation~\cite{zhou2023distillspec} of the target LLM.
However, in practice, an off-the-shelf draft model is not always available, requiring quantizing or pruning the original model, or even pretraining a smaller model from scratch.

To overcome the limitations of requiring a separate draft model, \emph{draft-model-free} SD methods have been proposed.
One common approach is to add auxiliary prediction layers on top of the final transformer block ~\cite{li2024eagle, li2024eagle2fasterinferencelanguage, li2025eagle3}.
These additional ``head'' layers take the hidden states as input and directly propose tokens.
However, practitioners still need to fine-tune these drafting heads separately.
Recently, a new line of work has removed the need for post-hoc training. Multi-Token Prediction (MTP)~\cite{deepseekai2025deepseekv3technicalreport,5team2025glm45agenticreasoningcoding} co-trains auxiliary prediction heads jointly with the main model, rather than fine-tuning them afterward. As a result, these auxiliary heads are available out of the box and achieve substantially higher token-acceptance rates.

Another line of draft-model-free SD methods utilize a non-LLM proposer.
A representative example is prompt lookup decoding, or \emph{$n$-gram}~\cite{saxena2023prompt}.
The intuition is that LLM outputs often contain recurring phrases; therefore, we can extract $n$-gram snippets from previously generated text and reuse them as token proposals.
This approach is particularly effective in workloads such as code or passage editing, where the input and output share substantial overlap. 
Another example is using retrieval-augmented generation (RAG) to retrieve candidate proposals from an external corpus~\cite{he2024restretrievalbasedspeculativedecoding}.
These non-LLM approaches demonstrate the potential for lightweight or task-specific drafting mechanisms.

\textbf{Problems of Existing Benchmarks.}
Most prior works evaluate SD by building proof-of-concept research prototypes, as integration into a production-grade inference system is nontrivial and requires modifications across multiple components.
However, this makes a fair comparison between SD methods difficult and the benefits in real-world deployment impossible to predict, due to the following limitations.
First, many prototype implementations lack key optimizations in production systems, such as the use of CUDA graphs~\cite{cudagraph} to reduce system overhead, a common feature in LLM serving systems~\cite{kwon2023efficient, zheng2024sglang}.
Second, the implementations are typically evaluated with a batch size of one, which is unrepresentative of large-scale batched deployments of the real world.
Finally, prior work lacks an in-depth analysis of SD speedup, as they primarily focus on high-level metrics such as average latency or dataset-level acceptance rate. \cite{li2024eagle,li2025eagle3,xia-etal-2024-unlocking}
We believe that a deeper understanding, such as execution time breakdown and position-level acceptance behavior, is necessary to fully explain and optimize the end-to-end performance of SD.
\section{End-to-End Performance of Speculative Decoding}
\label{sect:e2e-perf}
In this section, we evaluate the performance of SD variants across various model sizes and datasets.
The complete experimental setup is summarized in  \autoref{tab:exp-config} of the Appendix.

\subsection{Experiment Setup}
\textbf{Inference Engine and Hardware.}
All experiments are conducted using vLLM v0.10.1.1 unless otherwise stated.
To closely approximate real-world serving scenarios, we enable all default vLLM optimizations, 
including KV cache management, continuous batching, chunked prefill, and CUDA Graphs.
We run all experiments on NVIDIA H100~\cite{h100} with 80 GB of memory.
For 8B models, we use a single GPU, while for the 70B/106B models, we use four GPUs with a tensor parallel size of four. 

\textbf{Models and SD variants}
We evaluate two models from the Llama-3 family \cite{dubey2024llama}:
Llama3.1-8B-Instruct and Llama3-70B-Instruct.
We also include Qwen3-8B\footnote{Throughout the paper, Qwen3-8B-Thinking denotes the same model with thinking mode enabled, whereas Qwen3-8B refers to its non-thinking mode.}~\cite{yang2025qwen3technicalreport} and GLM-4.5-Air-106B \cite{5team2025glm45agenticreasoningcoding} for reasoning workloads. We benchmark the following SD variants:
\begin{itemize} [leftmargin=1.5em, itemsep=0pt]
    \item \textbf{Draft-model-based}~\cite{leviathan2023fast, chen2023accelerating}: Uses a smaller draft model to propose 3 draft tokens per step, which are later verified by the target model. Details of model pairing are in Appendix~\autoref{appendix:model-config}.
  \item \textbf{EAGLE}~\cite{li2024eagle} and \textbf{EAGLE-3}~\cite{li2025eagle3}: Draft-model-free methods with fine-tuned auxiliary prediction heads, predicting 3 draft tokens per step for chain-based settings\footnote{Official EAGLE-3 weights are unavailable for Llama-3-70B, EAGLE weights are likewise unavailable for Qwen3-8B and EAGLE-3/EAGLE weights are also unavailable for GLM-4.5-Air~\cite{li2024eagle}.}.
  \item \textbf{Multi-Token Prediction} (MTP)~\cite{deepseekai2025deepseekv3technicalreport}: Co-trained draft-model-free approach where auxiliary heads are jointly trained with the target model; we use 3 draft tokens per step.
  \item \textbf{$n$-gram}~\cite{saxena2023prompt, somasundaram2024pld+}: Training-free method reusing recurring $n$-gram phrases from the prompt; we use 3 draft tokens per step with {\tt prompt\_lookup\_max=7} and {\tt prompt\_lookup\_min=3}. For \texttt{InstructCoder} with Llama3.1-8B, Llama3-70B and Qwen3-8B without reasoning, we additionally evaluate $n$-gram with 5 draft tokens per step to study the effect of a larger proposal length.
  
\end{itemize}

Unless otherwise specified, the maximum generation length is capped at 8192 tokens.
For reasoning workloads AIME22-24 and GPQA-Main, we extend this limit to 32,768 tokens to prevent early truncation.
Decoding is performed with temperature set to 0 and no sampling truncation (\texttt{top\_p=1}, \texttt{top\_k=-1}).

\textbf{Workloads.}
We evaluate six datasets representing diverse real-world applications:
\textit{CNN/DailyMail}~\cite{hermann2015teaching, see-etal-2017-get} (summarization),
\textit{ShareGPT}~\cite{sharegpt} (multi-turn chat),
\textit{InstructCoder}~\cite{li2024instructcoder} (code editting),
\textit{GSM8K}~\cite{cobbe2021gsm8k} (grade-school math),
and two complex reasoning datasets: \textit{AIME22-24}~\cite{AIME-AoPS, AIME-HF} and \textit{GPQA-Main}~\cite{rein2023gpqagraduatelevelgoogleproofqa}. We detailed the dataset information in Appendix \autoref{appendix:dataset}. 




\textbf{Generation Length and Evaluation Metrics.}
Although in theory SD should preserve the same token distribution as standard decoding, in practice we observe discrepancies in the generated outputs—-{\em even under greedy decoding.} \autoref{tab:gen-len} illustrates this by showing variations in generation length when using the same set of prompts while changing the batch size and SD variant.
For each configuration, we replicate each request multiple times to match the target batch size.
We repeat the process for 500 requests and record the generation length for each request.
\begin{table}[h]
\centering
\caption{Average generation length (mean $\pm$ standard deviation) of the baseline (w/o SD) and different SD variants on the ShareGPT dataset using Llama3.1-8B across varying batch sizes.}
\label{tab:gen-len}
\resizebox{0.9\linewidth}{!}{
\begin{tabular}{ccccc}
\hline
\textbf{Batch Size} & \textbf{w/o SD} & \textbf{ngram} & \textbf{Eagle} & \textbf{Eagle3} \\
\hline
1   & 737$\pm$1048 & 750$\pm$1100 & 731$\pm$1010 & 736$\pm$1046 \\
8   & 722$\pm$994  & 676$\pm$810  & 709$\pm$928  & 704$\pm$916  \\
16  & 708$\pm$932  & 675$\pm$818  & 686$\pm$853  & 686$\pm$853  \\
32  & 674$\pm$784  & 697$\pm$887  & 693$\pm$857  & 706$\pm$917  \\
64  & 687$\pm$852  & 696$\pm$911  & 709$\pm$924  & 709$\pm$924  \\
128 & 714$\pm$930  & 739$\pm$1050 & 709$\pm$924  & 709$\pm$924  \\
\hline
\end{tabular}
}
\end{table}

As shown in \autoref{tab:gen-len}, even without speculative decoding (the w/o SD column), generation length differs across different batch sizes, and similarly when SD is used. Importantly, these differences are not consistently longer or shorter than the baseline, suggesting that they stem from numerical nondeterminism rather than implementation bugs.
Recent work~\cite{he2025nondeterminism} attributes such variability to kernel-level nondeterminism and floating-point variation during GPU execution.

To account for fluctuations in output length, we use token \textbf{throughput}, i.e., the number of generated tokens per second, as the performance metric.
This measure removes the confounding effect of varying generation lengths. \textbf{Speedup} is then defined as the ratio between the throughput with speculative decoding (SD) and the throughput without it.


\subsection{Results}
\begin{figure*}[h!]
\begin{center}
\captionsetup{justification=centering}
\includegraphics[width=0.6\linewidth]{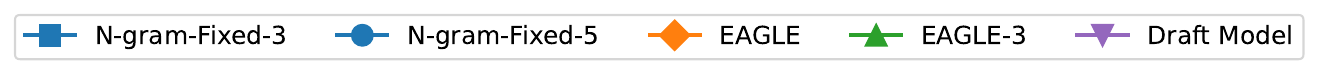}
  
\begin{subfigure}[t]{0.22\textwidth}
\includegraphics[width=\linewidth]{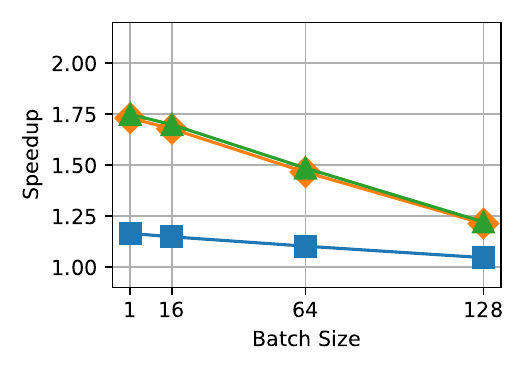}
\caption{Llama3.1-8B, GSM8K}
\end{subfigure}
\begin{subfigure}[t]{0.22\textwidth}
\includegraphics[width=\linewidth]{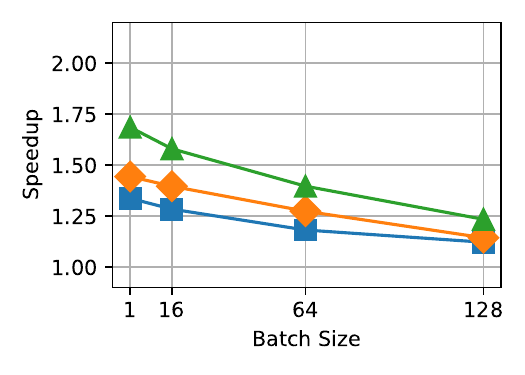}
\caption{Llama3.1-8B, CNN/Dailymail}
\end{subfigure}
\begin{subfigure}[t]{0.22\textwidth}
\includegraphics[width=\linewidth]{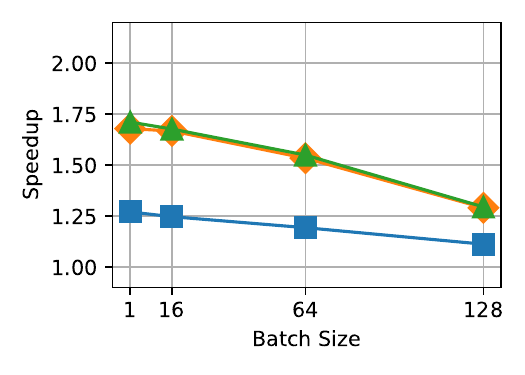}
\caption{Llama3.1-8B, ShareGPT}
\end{subfigure}
\begin{subfigure}[t]{0.22\textwidth}
\includegraphics[width=\linewidth]{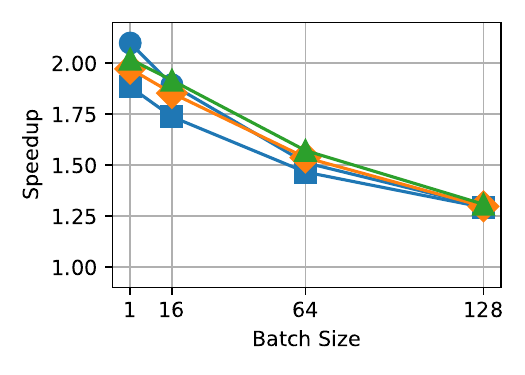}
\caption{Llama3.1-8B, InstructCoder}
\end{subfigure}

\begin{subfigure}[t]{0.22\textwidth}
\includegraphics[width=\linewidth]{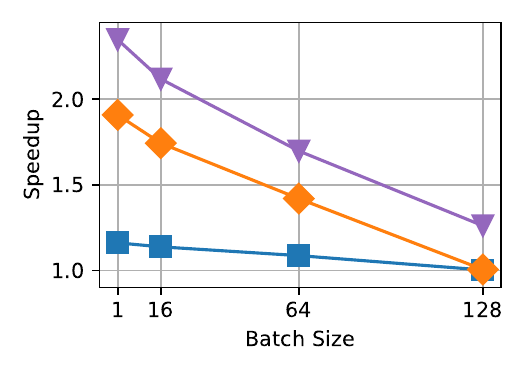}
\caption{Llama3-70B, GSM8K}
\end{subfigure}
\begin{subfigure}[t]{0.22\textwidth}
\includegraphics[width=\linewidth]{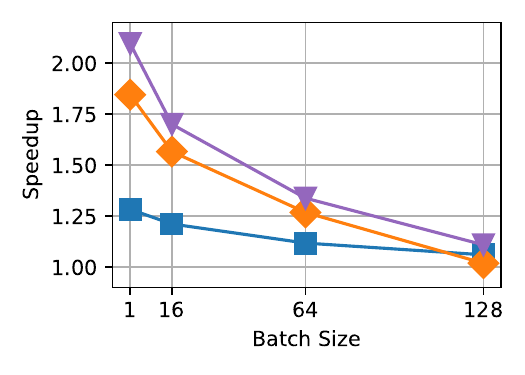}
\caption{Llama3-70B, CNN/Dailymail}
\end{subfigure}
\begin{subfigure}[t]{0.22\textwidth}
\includegraphics[width=\linewidth]{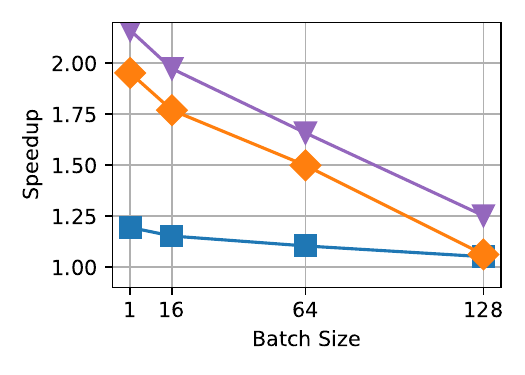}
\caption{Llama3-70B, ShareGPT}
\end{subfigure}
\begin{subfigure}[t]{0.22\textwidth}
\includegraphics[width=\linewidth]{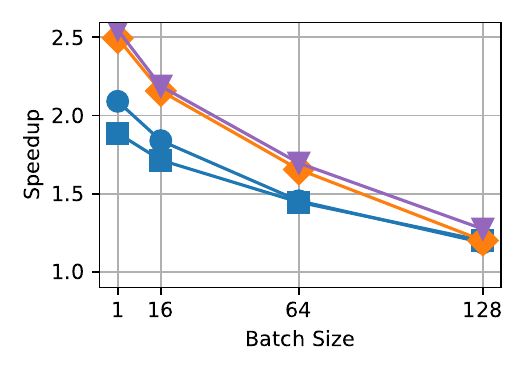}
\caption{Llama3-70B, InstructCoder}
\end{subfigure}

\begin{subfigure}[t]{0.22\textwidth}
\includegraphics[width=\linewidth]{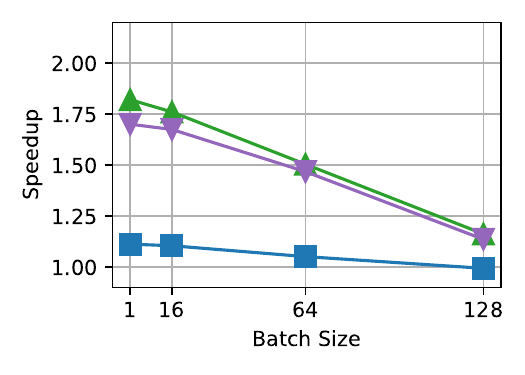}
\caption{Qwen3-8B, GSM8K}
\end{subfigure}
\begin{subfigure}[t]{0.22\textwidth}
\includegraphics[width=\linewidth]{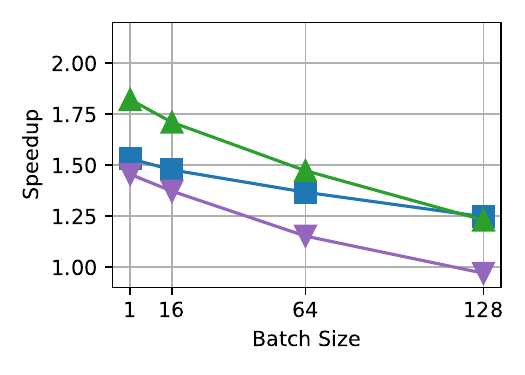}
\caption{Qwen3-8B, CNN/Dailymail}
\end{subfigure}
\begin{subfigure}[t]{0.22\textwidth}
\includegraphics[width=\linewidth]{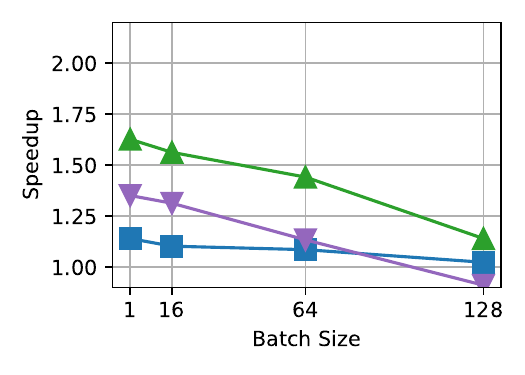}
\caption{Qwen3-8B, ShareGPT}
\end{subfigure}
\begin{subfigure}[t]{0.22\textwidth}
\includegraphics[width=\linewidth]{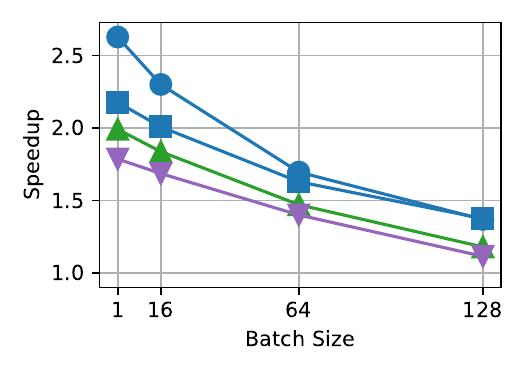}
\caption{Qwen3-8B, InstructCoder}
\end{subfigure}

\caption{End-to-end performance on non-reasoning workloads. For the $n$-gram method, we report results for both three-token and five-token proposals.}
\label{fig:e2e-speedup}
\end{center}
\end{figure*}

\textbf{End-to-end speedup.} We first present the end-to-end throughput comparison in \autoref{fig:e2e-speedup}.
Across most configurations, SD improves throughput over the baseline without SD. The improvement is most pronounced with small/medium
batch sizes, when the system is memory-bound, so lots of compute can be used for proposing and verification.

\begin{quoting}[leftmargin=1.5em, rightmargin=1.5em, vskip=0pt]
\textbf{Batch Size.}
\emph{Increasing batch size improves absolute throughput but systematically reduces the relative speedup of speculative decoding, an effect that is amplified for larger models.}
\end{quoting}

Consistent with prior work~\cite{liu2024dsd}, larger batch sizes yield higher absolute throughput but lower relative speedup. For example, for the Llama3.1-8B model on the GSM8K dataset, the speedup of EAGLE over the non-SD baseline decreases from 1.73$\times$ to 1.21$\times$ as the batch size increases from 1 to 128.

At a high level, SD trades additional computation for increased throughput or reduced latency. When the batch size is small, the system has sufficient idle compute resources, so the extra computation-spent on proposing and verifying tokens that are ultimately rejected—has limited impact. However, as the batch size grows, the system becomes increasingly compute-bound, making the overhead spent on proposing and verifying rejected tokens more costly, and thus leading to smaller speedups. As shown in \autoref{sec:breakdown}, the verification stage dominates the overall execution time under such conditions.

We further observe that this trade-off between computation and speedup becomes more obvious as the model size increases. For instance, with EAGLE, on the ShareGPT dataset, increasing the batch size from 1 to 32 reduces the speedup by 4.3\% (1.68$\times$ to 1.61$\times$) for Llama3.1-8B, whereas for Llama3-70B, the speedup drops by 14.0\% (1.96$\times$ to 1.72$\times$). This trend arises because the 70B model is executed on four GPUs, so even with small or medium batch sizes, the system is already compute-bound, leaving limited spare compute to verify tokens that are ultimately rejected.

\begin{quoting}[leftmargin=1.5em, rightmargin=1.5em, vskip=0pt]
\textbf{SD Variants and dataset.}
\emph{
The $n$-gram method is generally less effective than other SD approaches across most workloads, with the notable exception of code-editing tasks. Moreover, the draft-model-based method achieves the best performance on the 70B target model; however, the effectiveness diminishes as the target model size decreases (e.g., 8B).}
\end{quoting}
GSM8K, CNN/DailyMail, and ShareGPT all show similar trends, with EAGLE-3 or draft-model-based method achieving the highest speedup across batch sizes. 
In contrast, on InstructCoder, the performance gap among SD variants narrows. For both Llama3.1-8B and Qwen3-8B, the $n$-gram method even outperforms EAGLE and EAGLE-3. This behavior arises because code-editing workloads such as InstructCoder exhibit strong token reuse, which can be effectively exploited by simple drafting methods like $n$-gram.
We further analyze the performance of $n$-gram in \autoref{subsect:ngram-bleu-score}.


Next, we observe that draft-model–based methods outperform EAGLE on the Llama3-70B model in most settings. As shown in \autoref{fig:acc-len-request-level-box-plot}, draft-model–based approaches achieve higher token acceptance rates, while their proposing overhead is comparable to that of EAGLE (see \autoref{fig:exec-breakdown}). This behavior is expected: draft models are pretrained on large-scale corpora, whereas the EAGLE head is only fine-tuned on a relatively limited dataset, which naturally constrains its acceptance rate.

In contrast, on Qwen3-8B, draft-model–based speculative decoding consistently underperforms EAGLE-3 and, in some cases, even the training-free $n$-gram method.
Specifically, for Llama3-70B, we use Llama3.2-1B as the draft model, whereas for Qwen3-8B we use Qwen3-0.6B. Although acceptance rates can be similar across the 70B and 8B setups on the same dataset—or even higher for the 8B case sometimes—draft-model–based SD is nonetheless noticeably less effective on the smaller model. For example, on GSM8K, the acceptance rate is 75\% for the 70B setup and 81\% for the 8B setup, yet the overall performance gain is substantially lower for Qwen3-8B.

This discrepancy arises because the proposing overhead is substantially larger for the 8B model. Specifically, the execution-time ratio between the draft and target model's single forward pass is around 12.5\% for the 70B setup, compared to an estimated 37.5\% for the 8B setup. As a result, drafting becomes significantly more expensive relative to verification on smaller models, which diminishes the overall benefit of draft-model-based SD.




\begin{figure*}[h!]
\begin{center}
\captionsetup{justification=centering}
\includegraphics[width=0.8\linewidth]{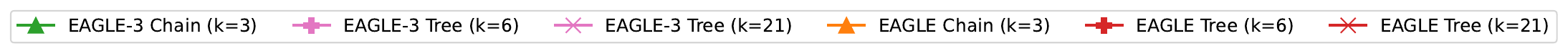}

\begin{subfigure}[t]{0.22\textwidth}
\includegraphics[width=\linewidth]{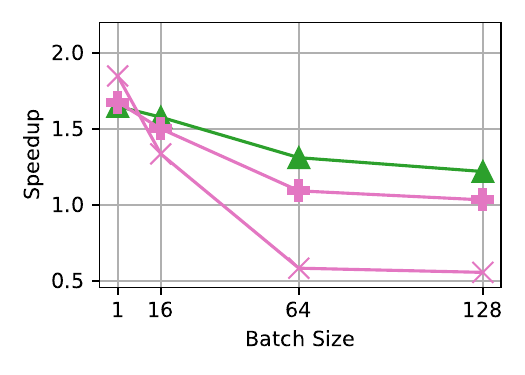}
\caption{Qwen3-8B, GSM8K}
\end{subfigure}
\begin{subfigure}[t]{0.22\textwidth}
\includegraphics[width=\linewidth]{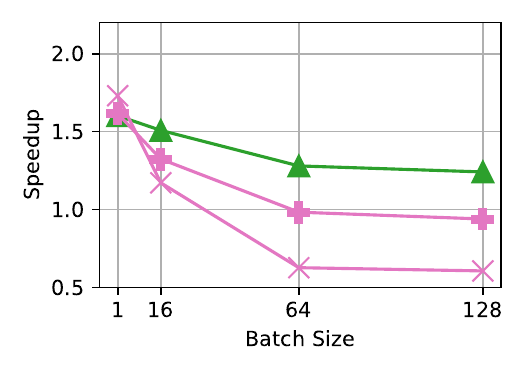}
\caption{Qwen3-8B, CNN/Dailymail}
\end{subfigure}
\begin{subfigure}[t]{0.22\textwidth}
\includegraphics[width=\linewidth]{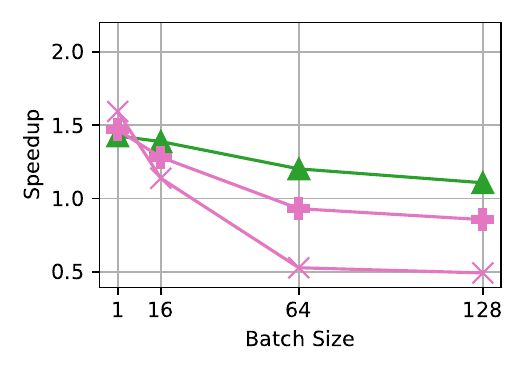}
\caption{Qwen3-8B, ShareGPT}
\end{subfigure}
\begin{subfigure}[t]{0.22\textwidth}
\includegraphics[width=\linewidth]{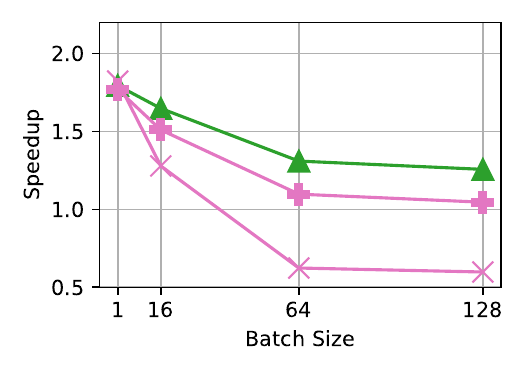}
\caption{Qwen3-8B, InstructCoder}
\end{subfigure}

\begin{subfigure}[t]{0.22\textwidth}
\includegraphics[width=\linewidth]{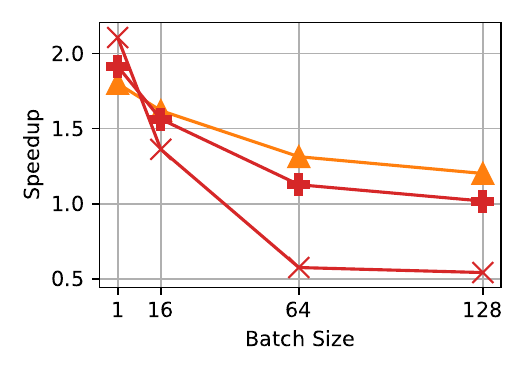}
\caption{Llama3-70B, GSM8K}
\end{subfigure}
\begin{subfigure}[t]{0.22\textwidth}
\includegraphics[width=\linewidth]{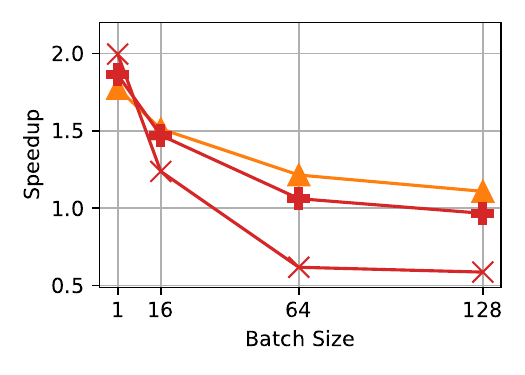}
\caption{Llama3-70B, CNN/Dailymail}
\end{subfigure}
\begin{subfigure}[t]{0.22\textwidth}
\includegraphics[width=\linewidth]{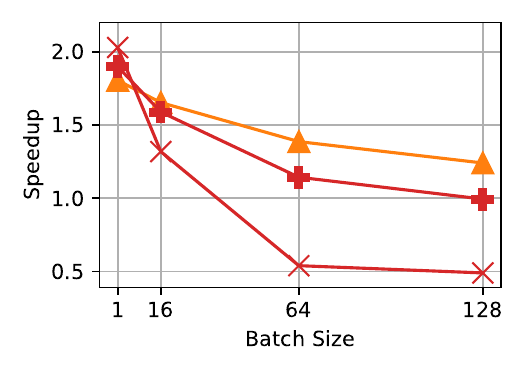}
\caption{Llama3-70B, ShareGPT}
\end{subfigure}
\begin{subfigure}[t]{0.22\textwidth}
\includegraphics[width=\linewidth]{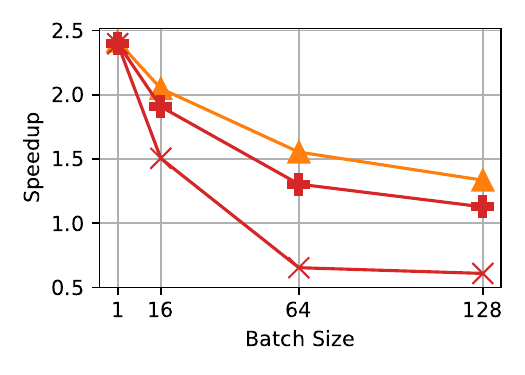}
\caption{Llama3-70B, InstructCoder}
\end{subfigure}
\caption{End-to-end performance on non-reasoning workloads for tree-style verification. We compare a chain setting ($k{=}3$) against tree settings with $k{=}6$ and $k{=}21$, using a fixed depth of 3 for all runs.}
\label{fig:e2e-speedup-tree}
\end{center}
\end{figure*}

\begin{quoting}[leftmargin=1.5em, rightmargin=1.5em, vskip=0pt]
\textbf{Tree-Style Verification.}
\emph{Tree-based methods provide slightly higher speedup at batch size 1, but the advantage quickly disappears as batch size increases. Across batch sizes, the chain-style method often remains the more performant.}
\end{quoting}


We next evaluate tree-style verification in SGLang v0.5.9~\cite{zheng2024sglang} using Qwen3-8B and Llama3-70B on the non-reasoning workloads. We use SGLang for this study because the draft-tree path in the vLLM version we evaluated is not yet sufficiently optimized for a fair comparison across tree configurations. To control the comparison, we fix the tree depth to 3 in all runs. We use chain-style verification with $k{=}3$ as the baseline, evaluate a wider tree with $k{=}21$ and branching factor 4 to approximately match the tree structure used in EAGLE~\cite{li2024eagle}, and include an intermediate tree with $k{=}6$ and branching factor 2. Here, $k$ denotes the number of draft tokens verified in parallel by the target model. Unless otherwise noted, all other settings follow those described earlier. By default, SGLang uses a dynamic draft-tree policy which optimises the tree structure during decoding~\cite{li2024eagle}. In this setup, FlashAttention-3~\cite{shah2024flashattention} is used for the chain configuration, whereas FlashInfer~\cite{ye2025flashinfer} is used for the tree configurations.

As shown in \autoref{fig:e2e-speedup-tree}, tree-based EAGLE/EAGLE-3 achieves slightly higher speedup than the chain-based settings at batch size 1. For example, on Qwen3-8B with GSM8K, speedup increases from 1.65$\times$ for the chain to 1.68$\times$ for $k{=}6$ and 1.85$\times$ for $k{=}21$. On Llama3-70B with ShareGPT, it rises from 1.81$\times$ to 1.90$\times$ and 2.03$\times$. However, this benefit quickly disappears as batch size increases. By batch size 64, the $k{=}21$ tree falls below 1$\times$ speedup on all workloads for both models, whereas the chain remains above 1$\times$ throughout.

This is likely because wider trees increase the accepted length, but also verify many more tokens that are later rejected. On Qwen3-8B with GSM8K, the accepted length increases from 2.25 for the chain to 2.51 and 2.92 for the $k{=}6$ and $k{=}21$ trees, respectively, while the acceptance rate decreases from 0.415 to 0.300 and 0.095. We observe the same trend on Llama3-70B with ShareGPT (accepted length: 2.29 $\rightarrow$ 2.55 $\rightarrow$ 2.93; acceptance rate: 0.429 $\rightarrow$ 0.310 $\rightarrow$ 0.097). As discussed in \autoref{sec:breakdown}, verification dominates the execution cost, so the overhead of these additional rejected branches becomes increasingly expensive at larger batch sizes. Thus, tree-style verification provides only a narrow low-batch-size benefit, while chain-style verification remains the more robust setting overall.

\begin{figure*}[h!]
\captionsetup{justification=centering}
\begin{center}
\begin{subfigure}[t]{0.22\textwidth}
\includegraphics[width=\linewidth]{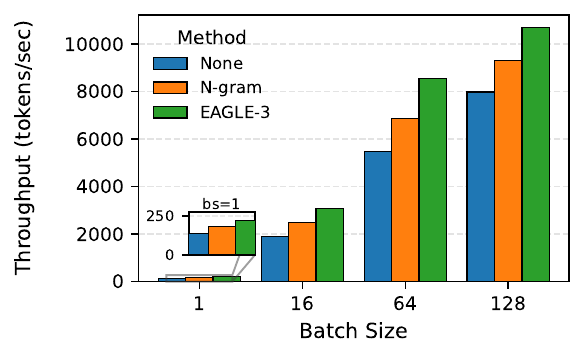}
\caption{Qwen3-8B-Thinking, GSM8K}
\end{subfigure}
\begin{subfigure}[t]{0.22\textwidth}
\includegraphics[width=\linewidth]{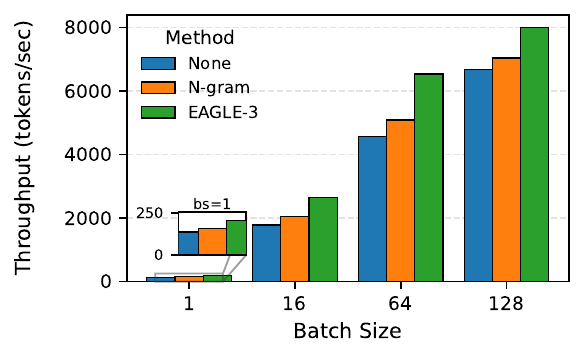}
\caption{Qwen3-8B-Thinking, CNN/Dailymail}
\end{subfigure}
\begin{subfigure}[t]{0.22\textwidth}
\includegraphics[width=\linewidth]{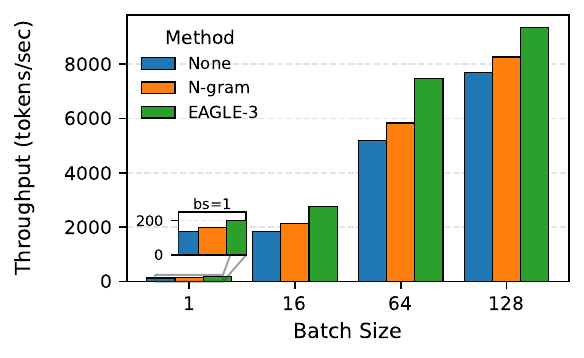}
\caption{Qwen3-8B-Thinking, ShareGPT}
\end{subfigure}
\begin{subfigure}[t]{0.22\textwidth}
\includegraphics[width=\linewidth]{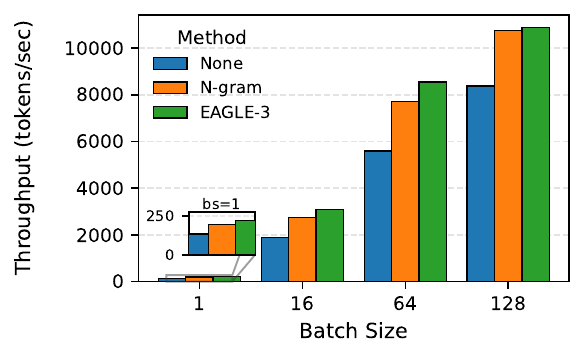}
\caption{Qwen3-8B-Thinking, InstructCoder}
\end{subfigure}

\begin{subfigure}[t]{0.22\textwidth}
\includegraphics[width=\linewidth]{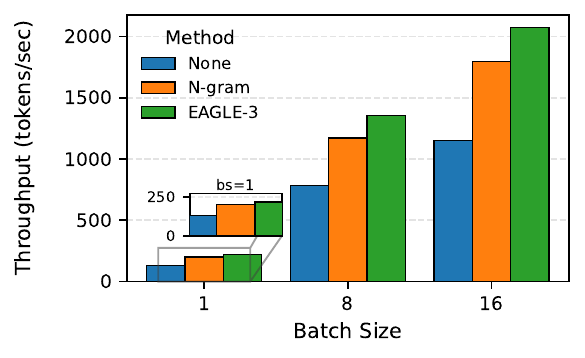}
\caption{Qwen3-8B-Thinking, AIME22-24}
\end{subfigure}
\begin{subfigure}[t]{0.22\textwidth}
\includegraphics[width=\linewidth]{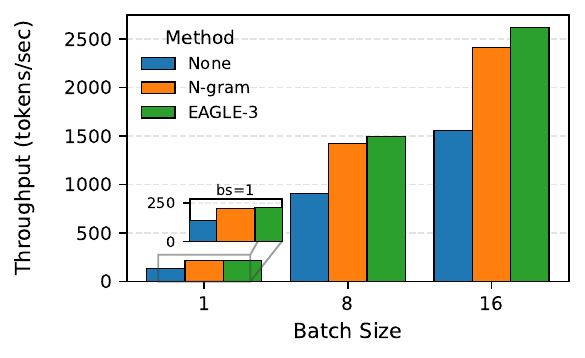}
\caption{Qwen3-8B-Thinking, GPQA-Main}
\end{subfigure}
\begin{subfigure}[t]{0.22\textwidth}
\includegraphics[width=\linewidth]{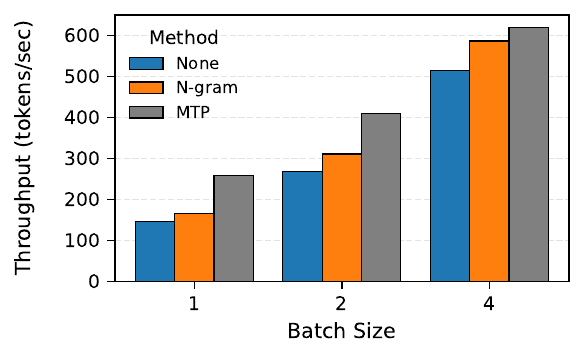}
\caption{GLM-4.5-Air, AIME22-24}
\end{subfigure}
\begin{subfigure}[t]{0.22\textwidth}
\includegraphics[width=\linewidth]{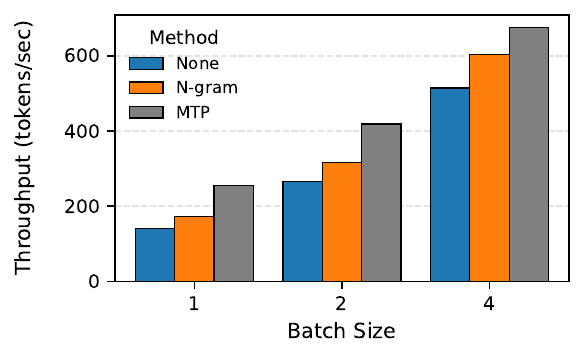}
\caption{GLM-4.5-Air, GPQA-Main}
\end{subfigure}
\caption{End-to-end performance on reasoning workloads. }
\label{fig:e2e-reasoning}
\end{center}
\end{figure*}

\begin{quoting}[leftmargin=1.5em, rightmargin=1.5em, vskip=0pt]
\textbf{Reasoning Workloads.}
\emph{Models and drafting methods that sustain high acceptance rates over long contexts—such as EAGLE-3 and, in some cases, $n$-gram—derive the greatest benefit on reasoning tasks. In contrast, MTP performance is limited by the reuse of a single MTP head across drafted tokens.}
\end{quoting}

As reasoning workloads become increasingly prevalent, understanding the effectiveness of speculative decoding (SD) in this regime becomes important. These workloads feature short prompts but long generation sequences. We evaluate SD on two representative reasoning benchmarks, AIME22–24~\cite{AIME-HF} and GPQA-Main~\cite{rein2023gpqagraduatelevelgoogleproofqa}.

As shown in \autoref{fig:e2e-reasoning}, we report the absolute generation throughput across different batch sizes. To ensure fair and consistent comparisons, we restrict our evaluation to medium batch sizes. Reasoning tasks typically involve long generation sequences, which can trigger request preemption once the KV cache becomes full. By avoiding such preemption, the reported results more accurately reflect the inherent efficiency of speculative decoding rather than performance artifacts caused by memory pressure.

For Qwen3-8B-Thinking, EAGLE-3 consistently leads across datasets, achieving 1.64$\times$--1.80$\times$ speedup over on GPQA-Main and AIME22–24, whereas $n$-gram performs comparably at 1.50$\times$--1.58$\times$. This trend likely arises because the model produces longer contexts and repetitive symbolic patterns, increasing opportunities for reuse. As shown later in~\autoref{sec:acc-behavior}, the average acceptance length in GPQA-Main generally grows with sequence length, with $n$-gram’s acceptance rising faster than EAGLE-3’s.

For GLM-4.5-Air, MTP outperforms $n$-gram but falls short of the theoretical upper bound, achieving 1.3$\times$--1.8$\times$ speedup on GPQA-Main. This gap is likely because the released open-source weights include only the first MTP module, which is trained primarily to predict the first token accurately but reused autoregressively for subsequent ones. Consequently, its position-wise acceptance rate (\(0.92 \rightarrow 0.68 \rightarrow 0.38\) on GPQA-Main), accuracy declines sharply across drafted tokens, shrinking the acceptance rate and thereby constraining the achievable speedup.

\section{Understanding the Performance}
\label{sec:understand-perf}
In this section, we aim to have a better understanding of the end-to-end speedups reported in \autoref{sect:e2e-perf}. 
We start by breaking down where time and memory are spent on different SD stages -- drafting, verification, rejection sampling, and others, and how these keep us from reaching the theoretical upper bound. We then examine how acceptance patterns within a request, across requests, and across datasets. 


\subsection{End to End Speedup Formula}
As pointed in the original SD paper~\cite{leviathan2023fast},
Let $c$ denote the execution-time ratio of a single forward pass between the drafting method and the target model, $\alpha$ is the token acceptance rate and $k$ is the proposed length. The expected overall wall-time speedup achieved by speculative decoding, can be expressed as follows:\footnote{This formulation assumes that the $k + 1$ simultaneous evaluations of the target model take approximately the same amount of time as generating a single token in parallel. This assumption typically holds when the system is memory-bound, which is common at small batch sizes, but may break down once the system becomes compute-bound.}
\begin{equation}
\label{eq:speedup}
     E(speedup)=\frac{1-\alpha^{k+1}}{(1-\alpha)(kc+1)}.
\end{equation}
As shown, the speedup depends on two factors: the proposed length ($k$), the execution-time ratio between the drafting method and the target model ($c$), and the token acceptance rate ($\alpha$). Importantly, the absolute execution time of the drafting method is irrelevant; only its runtime relative to the target model affects the achievable speedup.

Employing a more sophisticated drafting method can improve the token acceptance rate $\alpha$, but often at the cost of increased execution time, leading to a larger $c$. Consequently, optimal speedup is achieved by carefully choosing a drafting method that balances the trade-off between $c$ and $\alpha$. In the following sections, we evaluate these factors and additionally analyze the memory overhead of SD.

\section{Execution Time and Memory Breakdown}
\label{sec:breakdown}

\subsection{Execution Time}
\begin{figure*}
\centering
\begin{subfigure}[t]{0.43\linewidth}
\includegraphics[width=\linewidth]{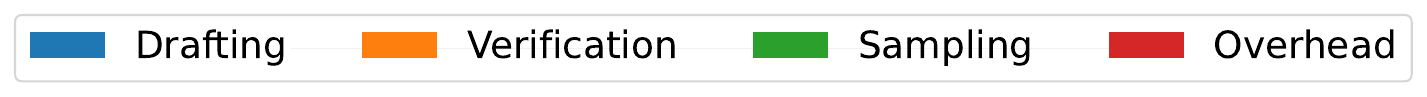}
\end{subfigure}

\begin{subfigure}[t]{0.33\linewidth}
\includegraphics[width=\linewidth]{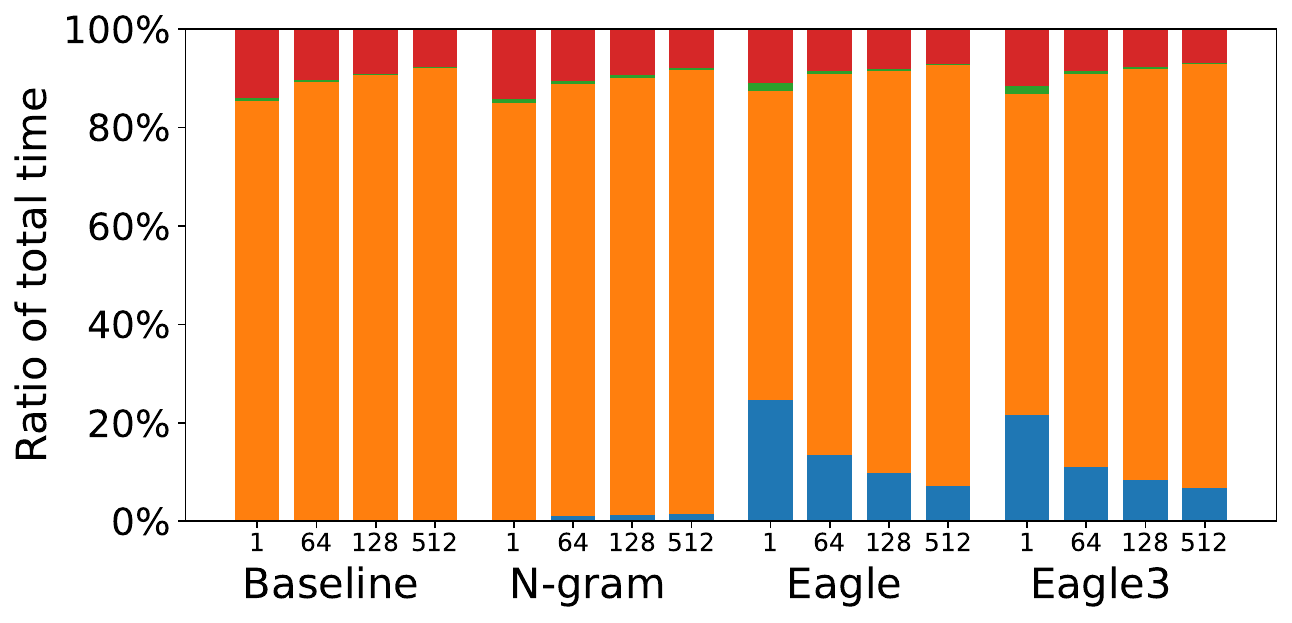}
\caption{Llama3.1-8B}
\end{subfigure}
\begin{subfigure}[t]{0.33\linewidth}
\includegraphics[width=\linewidth]{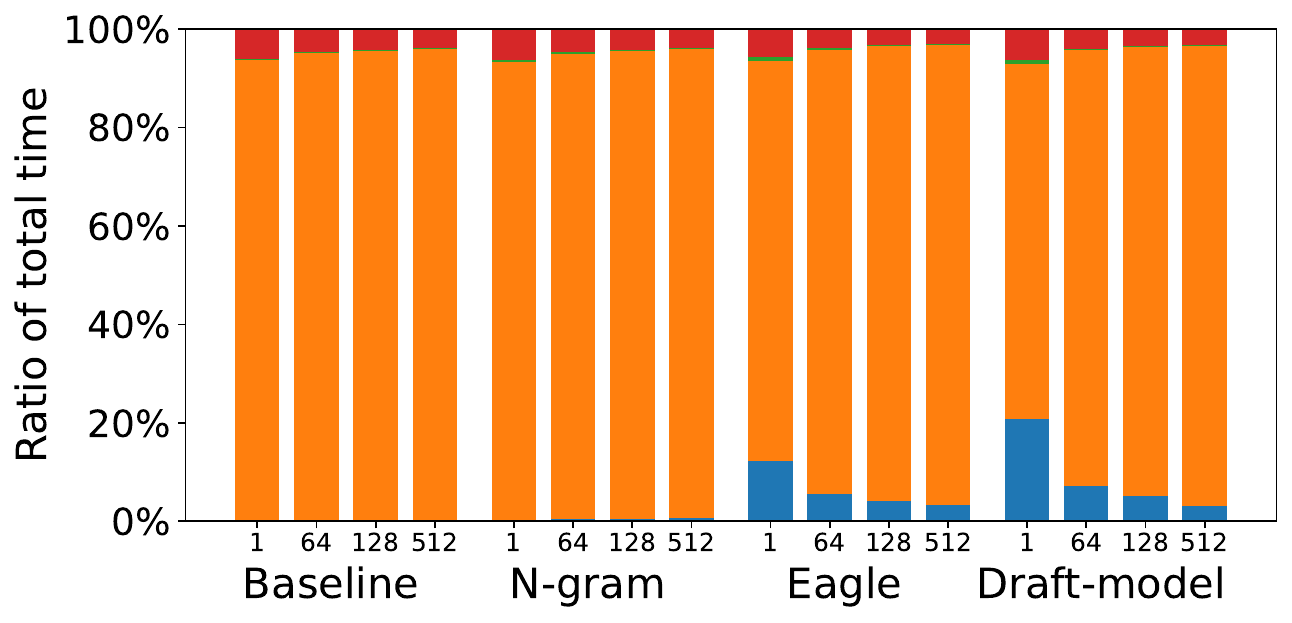}
\caption{Llama3-70B}
\end{subfigure}
\begin{subfigure}[t]{0.33\linewidth}
\includegraphics[width=\linewidth]{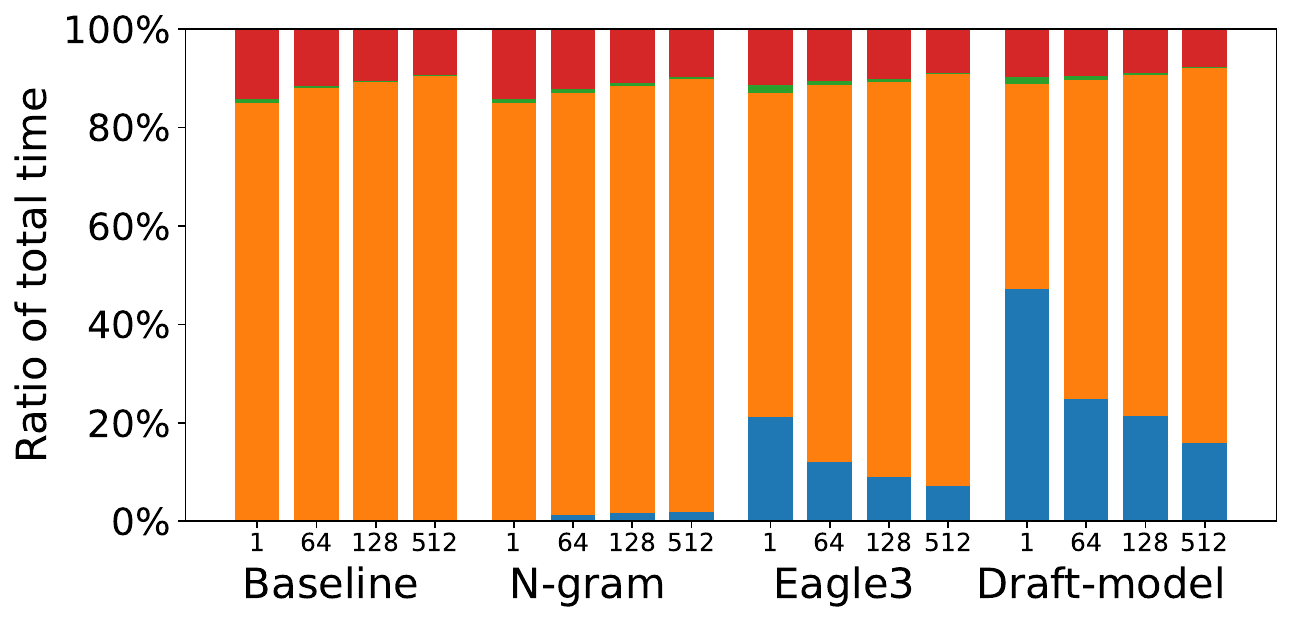}
\caption{Qwen3-8B}
\end{subfigure}
\caption{Execution time breakdown across different models.}
\label{fig:exec-breakdown}
\vspace{-0.1in}
\end{figure*}

In \autoref{fig:exec-breakdown}, we present the execution time breakdown of different stages in the baseline LLM execution and SD settings. For all experiments, we sample 500 requests from CNN/DailyMail. All other runtime settings are the same as \autoref{tab:exp-config}. 
For each configuration, we report the fraction of total execution time attributable to four components:
\textit{(i)}~\textbf{Drafting}, the time to generate speculated tokens, which corresponds to the lookup operation for $n$-gram and the autoregressive generation of the EAGLE heads for EAGLE and EAGLE-3;
\textit{(ii)}~\textbf{Verification}, the time spent by the target model to validate proposed tokens;
\textit{(iii)}~\textbf{Rejection Sampling}, time spent on generating final tokens based on verification logits;
and \textit{(iv)}~\textbf{Other Overheads}, all other overheads in the vLLM execution system. 

\begin{quoting}[leftmargin=1.5em, rightmargin=1.5em, vskip=0pt]
\textbf{Execution Time.}
\emph{Verification time generally accounts for the largest share of runtime across speculative decoding variants, while sampling time is negligible. Drafting is negligible for $n$-gram, stays below 20\% for EAGLE/EAGLE3, and stays under 40\% for draft-model methods acorss all batch sizes.}
\end{quoting}

We observe that the verification stage generally takes the largest execution time, ranging from 42\% to 95\% in all execution methods, which increases with the model size and batch size. 

Drafting costs, however, vary substantially by proposal mechanism. For $n$-gram, drafting time accounts for less than 2\% across all executions, adding negligible overhead to the baseline execution. Both EAGLE and EAGLE3 show similar drafting overhead characteristics, accounting for a 12\% to 20\% of the execution time at batch size 1, and 3\% to 7\% at batch size of 512.
Draft-model-based methods differ in that they must run a separate smaller model autoregressively to propose tokens, leading to a larger drafting overhead when the batch size is small. For Llama3-70B, drafting takes from 21\% at batch size of 1 to 3\% at batch size of 512. For Qwen3-8B, drafting is about 47\% at batch size of 1 and decreases to 16\% at batch size of 512. As a result, the verification fraction is correspondingly lower (roughly 72\% for Llama3-70B and 42\% for Qwen3-8B at batch size of 1). 

Furthermore, the sampling stage takes a negligible amount of time, accounting for less than 1.7\% of total time across all executions. 
Lastly, vLLM overhead accounts for 3–12\% of total execution time. This fraction decreases as model and batch size grow, since fixed overheads—such as scheduling—are independent of model scale and therefore become amortized at larger batch size or model size.

\textbf{Implication.} Verification cost dominates the end-to-end execution of speculative decoding, indicating that running the large target model remains the primary computational bottleneck. When the proposed tokens are ultimately accepted, this cost is well justified. However, verifying tokens that are later rejected can incur substantial wasted computation. This observation motivates a natural question: how much of the verification compute is truly useful? Conversely, what speedup could be achieved if verification were performed only on “correct” tokens, eliminating wasted work? These questions form the basis of \autoref{sec:theoretical-speedup}, where we estimate the theoretical upper bound for SD.

\subsection{Memory}
In \autoref{tab:memory}, we present the GPU memory usage of each SD variant with Llama3.1-8B, Llama3-70B and Qwen3-8B.
We use FP-16 precision and report the static memory overhead from the model weights and per-token KV cache calculated using model specs. We do not report intermediate memory overheads, such as the activation tensors during execution. The detailed calculation steps can be found in \autoref{appendix:mem}.

\begin{table}[t] 
    \small
    \begin{subtable}[t]{\columnwidth}
        \centering
        \begin{tabular}{c|c|c|c}
             \toprule
              & No SD/$n$-gram & EAGLE & EAGLE 3\\
             \midrule        
             Static (GiB) & 14.96 & 15.43 & 15.75 \\
             Per-token (KiB) & 128 & 132 & 132 \\
             \bottomrule
        \end{tabular}
        \caption{Memory breakdown on Llama3.1-8B} 
        \label{tab:sub1} 
    \end{subtable}
    \hfill 
    \begin{subtable}[t]{\columnwidth}
        \centering
        \begin{tabular}{c|c|c|c}
              \toprule
              & No SD/$n$-gram & EAGLE & D-M (1B)\\
             \midrule        
             Static (GiB) & 131.4 & 133.3 & 133.7 \\
             Per-token (KiB) & 320 & 324 & 352\\
             \bottomrule
        \end{tabular}
        \caption{Memory breakdown on Llama3-70B} 
        \label{tab:sub2} 
    \end{subtable}
    \hfill
    \begin{subtable}[t]{\columnwidth}
        \centering
        \begin{tabular}{c|c|c|c}
              \toprule
              & No SD/$n$-gram & EAGLE3 & D-M (0.6B)\\
             \midrule        
             Static (GiB) & 15.25 & 16.00 & 16.37 \\ 
             Per-token (KiB) & 144 & 148 & 256\\
             \bottomrule
        \end{tabular}
        \caption{Memory breakdown on Qwen3-8B} 
        \label{tab:sub3}
    \end{subtable}
    
    \caption{Static and per-token memory usage of the baseline model execution and different speculative decoding methods. DM stands for Draft-Model-based method.}
    \label{tab:memory} 
    \vspace{-0.1in}
\end{table}

\begin{quoting}[leftmargin=1.5em, rightmargin=1.5em, vskip=0pt]
\textbf{Memory.}
\emph{Speculative decoding generally incurs minimal memory overhead, including both static parameters and KV cache: it is zero for $n$-gram, below 10\% for EAGLE/EAGLE-3, while draft-model-based methods incur overhead that depends on the chosen draft model.}
\end{quoting}
We find that the memory overhead of speculative decoding (SD) is minimal. For $n$-gram SD, draft tokens are sampled from CPU-resident generation history, incurring no GPU memory overhead. 

\textbf{Static Memory.} Static memory corresponds to the memory required to store model weights. Certain SD variants add extra layers or even an independent model, thereby incurring additional static memory overhead. EAGLE-based SD introduces an additional Transformer layer and thus requires extra static weights, but the overhead is small: 3.1\% (EAGLE)/ 5.3\% (EAGLE3) for Llama-3.1-8B, 1.4\% (EAGLE) for Llama-3-70B, and 4.9\% (EAGLE3) for Qwen3-8B.
For draft-model-based SD, static memory overhead depends on the size of the draft model; in our setup, pairing Llama-3-70B with Llama-3.2-1B incurs a 1.8\% increase while pairing Qwen3-8B with Qwen3-0.6B incurs a more significant 7.3\% increase. 

\textbf{Per-token Memory.} Per-token memory refers to the size of the KV cache allocated for each generated token. The total KV cache footprint grows with the number of generated but unfinished requests. Consequently, for workloads with long generation lengths such as reasoning workloads, memory cost introduced by KV cache becomes a dominant factor in overall memory consumption.
Per-token overhead is modest for EAGLE-based methods, as each EAGLE layer requires the same KV cache as a single attention layer (4 KiB per token), resulting in a 3.1\% overhead for Llama-3.1-8B and 1.3\% for Llama-3-70B. 
In contrast, draft-model–based SD incurs substantially higher per-token memory overhead, as it relies on an additional multi-layer model for token proposal. In our configuration, pairing a 0.6B draft model with an 8B target model increases the per-token memory footprint from 144 KiB to 256 KiB, corresponding to a 1.77$\times$ increase.

\begin{figure}[h!]
  \centering
  \begin{subfigure}{0.45\linewidth}
    \includegraphics[width=\linewidth]{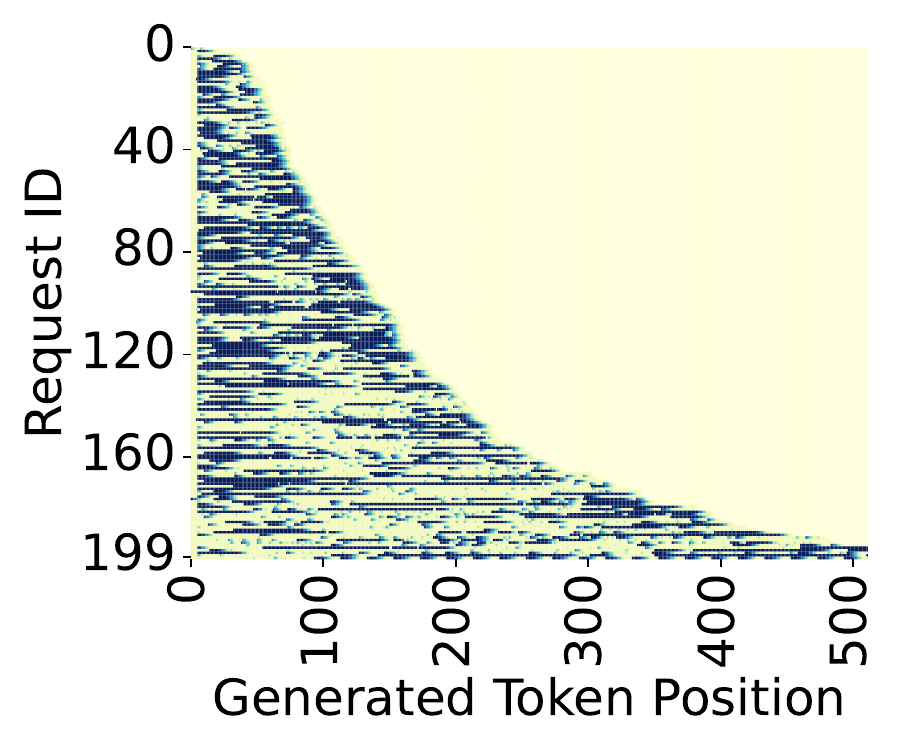}
    \caption{InstructCoder, $n$-gram}
  \end{subfigure}
  \hspace{1pt}
  \begin{subfigure}{0.45\linewidth}
    \includegraphics[width=\linewidth]{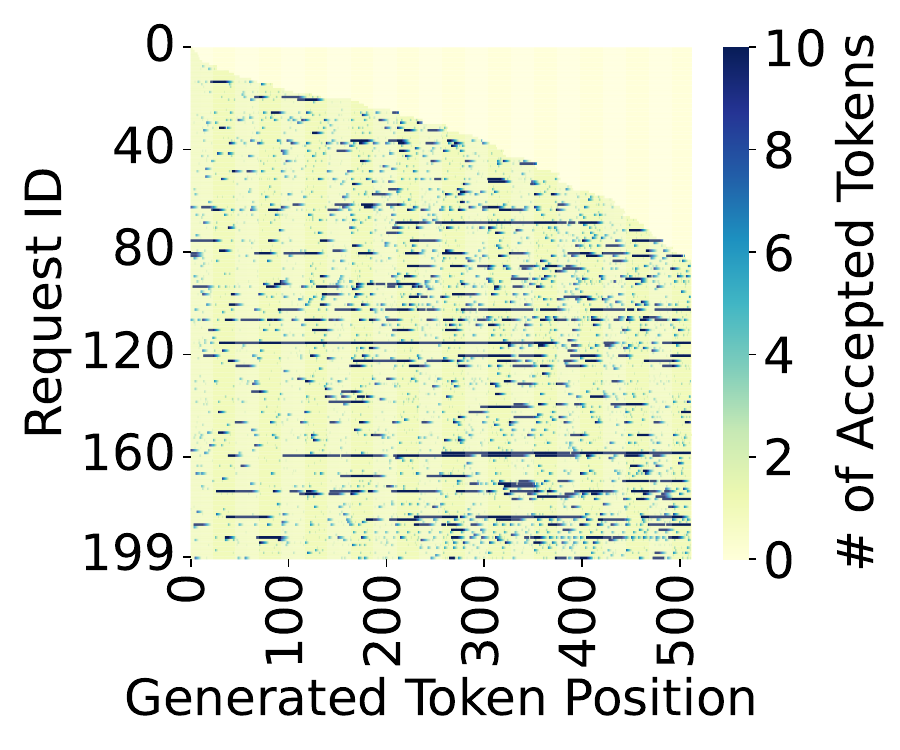}
    \caption{ShareGPT, $n$-gram}
  \end{subfigure}

   \begin{subfigure}{0.45\linewidth}
    \includegraphics[width=\linewidth]{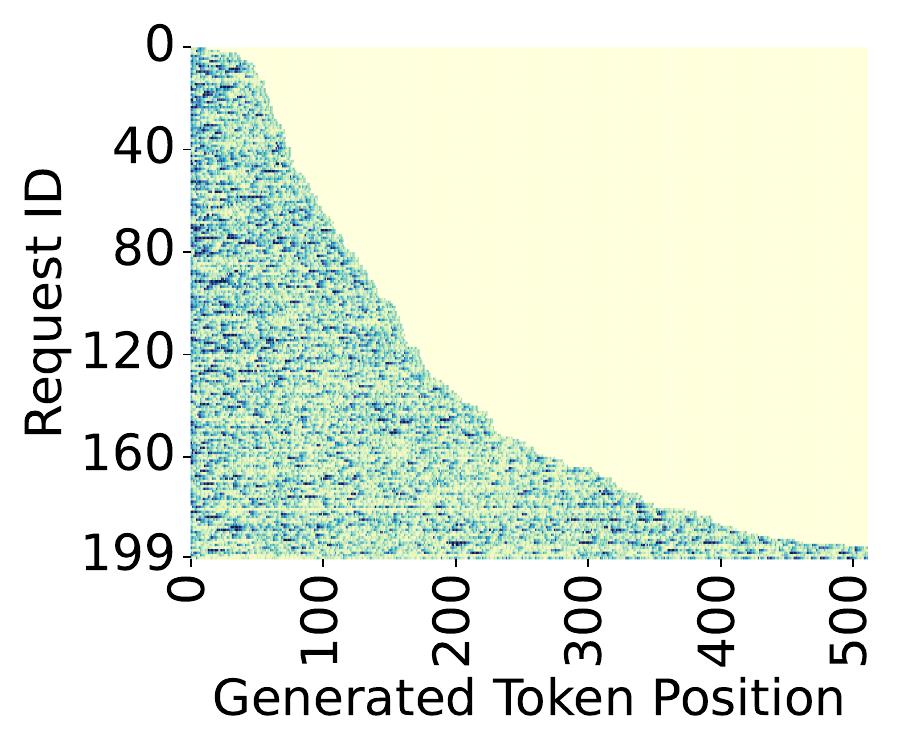}
    \caption{InstructCoder, EAGLE}
  \end{subfigure}
  \hspace{1pt}
  \begin{subfigure}{0.45\linewidth}
    \includegraphics[width=\linewidth]{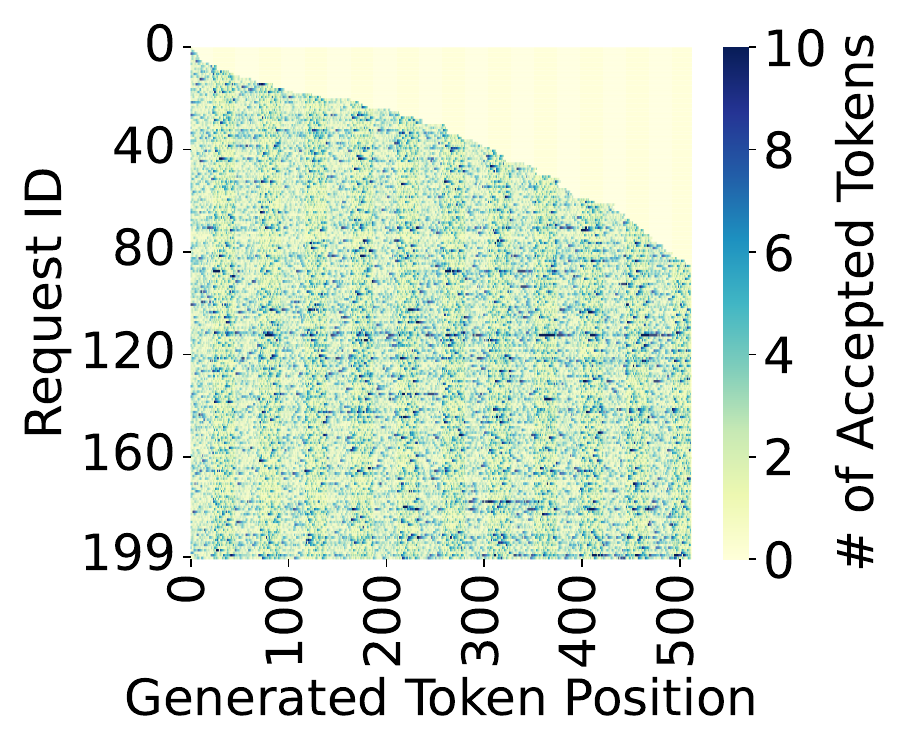}
    \caption{ShareGPT, EAGLE}
  \end{subfigure}

  \caption{Generation length per token position for Llama3.1-8B. Requests are sorted based on generation length. Darker colors indicate that more tokens are generated at the corresponding position.}
  \label{fig:acceptance-length-heatmap}
\end{figure}

\section{Acceptance Behavior}
\label{sec:acc-behavior}
To understand the acceptance rate, we measure the number of draft tokens generated\footnote{Number of generated tokens = accepted tokens proposed by the drafting method + 1 bonus token. Similarly, for generation length, it is computed by adding the acceptance length by one.} by the target model at each decoding step. 
For each dataset, we sample up to 200 requests. Each SD method proposes up to 20 tokens per step, and we record how many of these are generated at that step.  
After each generation step, even if multiple tokens were generated, we discard all but one and advance to the next position. 
Essentially, we approximate the maximum number of tokens that could be accepted at each generation step.
We analyze results for two non-reasoning models (Llama-3-70B and Llama-3.1-8B; max 512 output tokens) and one reasoning model (Qwen-3-8B-Thinking; max 32K output tokens).

\autoref{fig:acceptance-length-heatmap} visualizes the number of tokens generated at each output position for InstructCoder and ShareGPT (notice SD generates at least one token at each position, even if all proposed tokens are rejected). Each row represents a sampled request and each column a generation position, with color indicating the number of tokens produced by SD. 

\begin{quoting}[leftmargin=1.5em, rightmargin=1.5em, vskip=0pt]
\textbf{Summary.}
\emph{We observe the variance of generated token lengths along three dimensions: within a single request, across requests, and across datasets.}
\end{quoting}
\autoref{fig:acceptance-length-heatmap} gives a high level feeling of the generation length at each position across requests. $n$-grams and EAGLE show different patterns. $n$-grams show high variance, lots of very dark and very light positions, some positions generate a lot, while others only generate one. EAGLE show good average generation length (averge color is darker) and also less variance. But there are variance across all dimentions, which we will describe in more detail below.



\begin{quoting}[leftmargin=1.5em, rightmargin=1.5em, vskip=0pt]
\textbf{Within A Request.}
\emph{Longer generation workloads yield more accepted tokens per position, with $n$-gram benefiting from repetitive reasoning patterns but losing effectiveness near the end as the generation shifts toward conclusions.}

\end{quoting}

\begin{figure} [h!]
\captionsetup{justification=centering}
  \centering
  \begin{subfigure}{0.45\linewidth}
    \includegraphics[width=\linewidth]{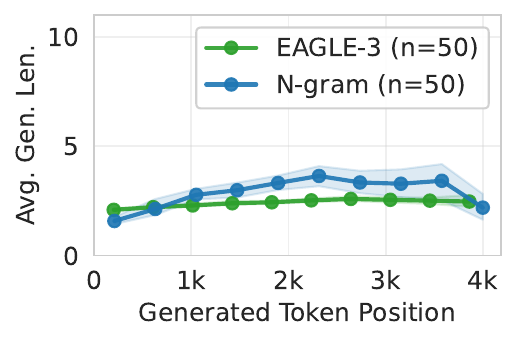}
    \caption{Requests with outputs shorter than 4K tokens}
  \end{subfigure}
    \hspace{1pt}
    \begin{subfigure}{0.45\linewidth}
    \includegraphics[width=\linewidth]{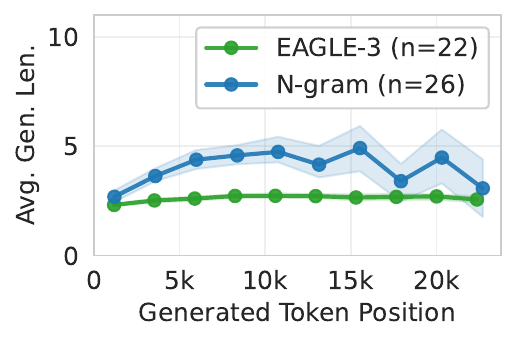}
    \caption{Requests with outputs longer than 13K tokens}
  \end{subfigure}
  \begin{subfigure}{0.45\linewidth}
    \includegraphics[width=\linewidth]{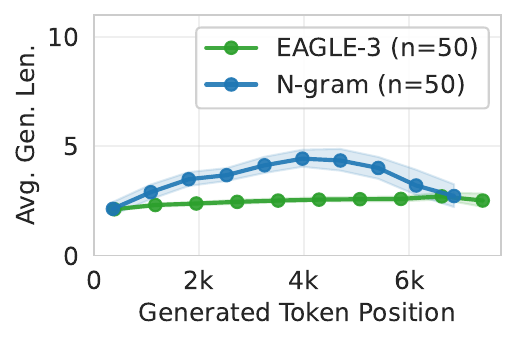}
    \caption{Requests with outputs of 4K-8K tokens}
  \end{subfigure}
    \hspace{1pt}
    \begin{subfigure}{0.45\linewidth}
    \includegraphics[width=\linewidth]{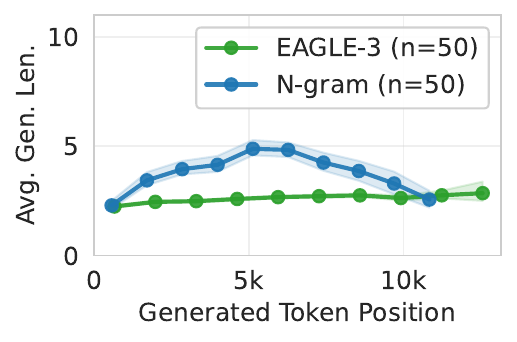}
    \caption{Requests with outputs of 8K-13K tokens}
  \end{subfigure}
  \vspace{-0.1in}
\caption{Average generation length (tokens) per output token position for Qwen3-8B-Thinking on GPQA-Main. Shaded bands show the 95\% confidence interval.}
  \label{fig:acc-len-token-level}
\end{figure}

Since different requests can have varying generation lengths, we sample 50 requests from each generation-length category: requests with generation length $<$4K, between 4–8K, between 8–13K, and over 13K. 
In \autoref{fig:acc-len-token-level}, token positions are partitioned into ten intervals, and the mean generation length is computed and plotted for each interval.

For reasoning workloads, generation length generally increases as generation progresses for both methods, but the growth is substantially steeper for $n$-gram. As shown in \autoref{fig:acc-len-token-level}, on GPQA-Main with Qwen-3-8B-Thinking, $n$-gram rises from roughly $1.6$--$3.7$ generated tokens across the sequence in short outputs ($<$4K tokens) to $2.7$--$5$ tokens in the longest responses ($>$13K tokens). 
EAGLE-3 grows much more gradually from around $2.1$--$2.5$ among the shortest requests and to roughly $2.5$--$5.7$ tokens among the longest requests. 
These results indicate that long scientific and technical generations accumulate locally repetitive structures, such as recurring variables, equations, and phrases, that $n$-gram can increasingly exploit. On the other hand, \textit{EAGLE}-based methods remain less sensitive to such surface-level recurrence.

For $n$-gram, we additionally observe a decrease towards the last few token position intervals. This is likely because the model shifts from step-by-step reasoning, where phrases and patterns tend to repeat a lot, to writing the final answer and stopping, which is less repetitive and therefore harder for $n$-gram to reuse previous context. Thus, in the last few token intervals, more requests are in the final-answer stage, so the average generation length drops.



\begin{figure} [t!]
  \centering
  \includegraphics[width=0.8\linewidth]{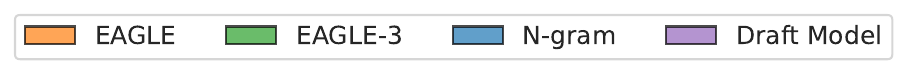}

  \begin{subfigure}{0.45\linewidth}
    \includegraphics[width=\linewidth]{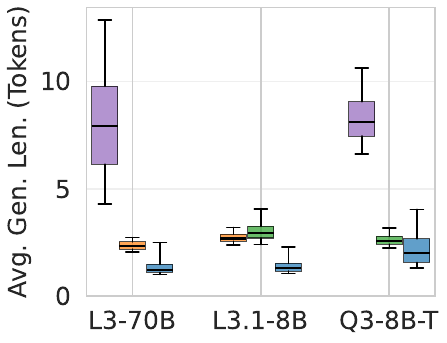}
    \caption{GSM8K}
  \end{subfigure}
      \hspace{1pt}
    \begin{subfigure}{0.45\linewidth}
    \includegraphics[width=\linewidth]{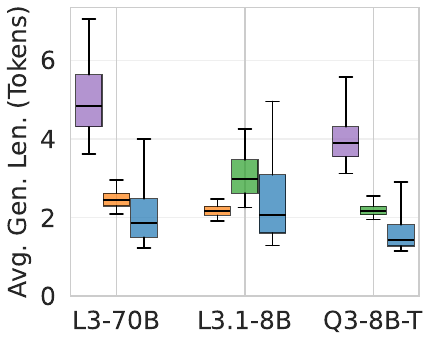}
    \caption{CNNDailyMail}
  \end{subfigure}
      \hspace{1pt}
  \begin{subfigure}{0.45\linewidth}
    \includegraphics[width=\linewidth]{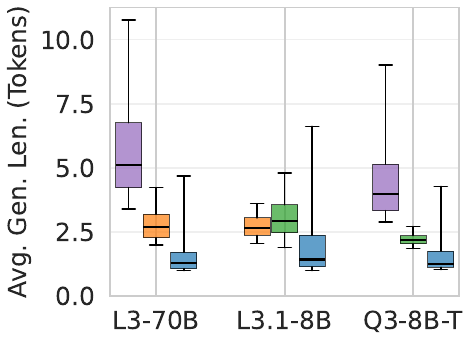}
    \caption{ShareGPT}
  \end{subfigure}
      \hspace{1pt}
  \begin{subfigure}{0.45\linewidth}
    \includegraphics[width=\linewidth]{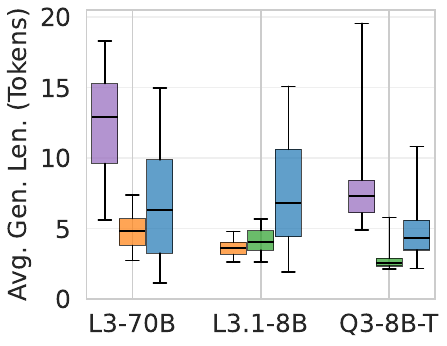}
    \caption{InstructCoder}
  \end{subfigure}
\caption{Request-level generation length across datasets and models. Each box shows the distribution of per-request mean generation length, with the box spanning the 25th–75th percentiles and whiskers covering the 5th–95th percentiles. Models from left to right are Llama3.1-8B, Llama3-70B and Qwen3-8B-Thinking. Results for reasoning datasets (AIME22–24 and GPQA-Main) are shown in~\autoref{fig:acc-len-request-level-box-plot-reasoning}.}
  \label{fig:acc-len-request-level-box-plot}
\end{figure}

\begin{quoting}[leftmargin=1.5em, rightmargin=1.5em, vskip=0pt]
\textbf{Across Requests and Datasets.}
\emph{Draft-model-based method typically achieves the longest median generation lengths across all workloads, EAGLE attains more stable per-request generation lengths, whereas $n$-gram shows higher variance and a heavy-tailed generation-length distribution.}
\end{quoting}

We observe substantial variance in per-request generation length in \autoref{fig:acc-len-request-level-box-plot}. Even within the same workload, the efficiency of speculative decoding can differ drastically. For example, on InstructCoder with Llama3-70B, EAGLE produces relatively short request-level generation lengths (2.7-7.4 tokens), whereas $n$-gram and draft-model-based SD achieve longer lengths with wider spreads (1.1–15.0 tokens and 5.6-18.3 tokens, respectively). In other workloads (e.g. ShareGPT and CNN/DailyMail) and target models (e.g. Qwen3-8B), draft-model-based methods similarly shows higher generation lengths than $n$-gram and EAGLE.


These distributions reveal distinct proposal behaviors for each SD variant. Draft-model-based methods often achieve substantially longest generation lengths, but their whiskers (5th-95th percentiles) are also broadly spread. 
The reason behind this may be request-dependent alignment: for some requests, the draft model can track the target model closely so that many proposed tokens are accepted, while for others, it diverges early and only a few draft tokens are accepted after verification.

For $n$-gram, it exhibits heavy tails and large variance across requests, with standard deviations roughly 2x-5x higher than those of EAGLE-based methods. 
While most $n$-gram requests produce short accepted spans, a small subset yield exceptionally long bursts (often exceeding 15 tokens), as seen as long blue strips in \autoref{fig:acceptance-length-heatmap}. 
In code-editing tasks, these typically correspond to locally repetitive content (e.g., reused identifiers, function templates, or class definitions). We select and show an example in \autoref{fig:ngram_prompt_output_full}. Such bursty acceptance behavior reveals that $n$-gram speculation relies on discrete pattern matches that may occur infrequently but can produce large payoffs when present. Consequently, $n$-gram's performance is high when the local contexts are repetitive, but it remains unstable across requests and workloads.

In contrast, EAGLE and EAGLE-3 show compact, symmetric distributions centered near their medians (typically 2–4 tokens), indicating steadier acceptance driven by learned contextual representations rather than surface-level overlap.

These distributional behaviors are consistent with the end-to-end outcomes in \autoref{fig:e2e-speedup}. Draft-model-based methods deliver the largest speedups when the cost of running a separate draft model is relatively low. EAGLE/EAGLE3 follows and provides modest gains on most workloads. $n$-gram achieves its best speedups on highly repetitive code-editing tasks, where rare but very long bursts occur. Outside such settings, its heavy-tailed and unstable acceptance makes performance less predictable.

\begin{figure*}[t]
\centering
\lstset{style=codebox, language=Python}

\begin{minipage}[t]{0.485\textwidth}
\textbf{Prompt}
\begin{lstlisting}
class Car:

    def __init__(self, make, model, year, color):
        self.make = make
        self.model = model
        self.year = year
        self.color = color


car1 = Car('Toyota', 'Camry', 2018, 'Red')
print(car1.color)

# Replace the string data type used for car color with an enum data type. 
# Just output the code, do not include any explanation.
\end{lstlisting}
\end{minipage}\hfill
\begin{minipage}[t]{0.485\textwidth}
\textbf{Output}
\begin{lstlisting}
from enum import Enum


class Color(Enum):
    RED = 1
    BLUE = 2
    GREEN = 3
    YELLOW = 4
    BLACK = 5
    WHITE = 6


class Car:

    def __init__(self, make, model, year, color):
        self.make = make
        self.model = model
        self.year = year
        self.color = color


car1 = Car('Toyota', 'Camry', 2018, Color.RED)
print(car1.color)
\end{lstlisting}
\end{minipage}

\vspace{-0.4em}
\caption{\textbf{Example of prompt-output repetition in code editing from InstructCoder.} The output preserves most of the prompt and applies a localized edit, a structure that $n$-gram can exploit to produce occasional long accepted spans.}
\label{fig:ngram_prompt_output_full}
\vspace{-1.0em}
\end{figure*}
\begin{figure*}[h!]
\begin{center}
\captionsetup{justification=centering}
\begin{subfigure}{0.24\linewidth}
\includegraphics[width=\linewidth]{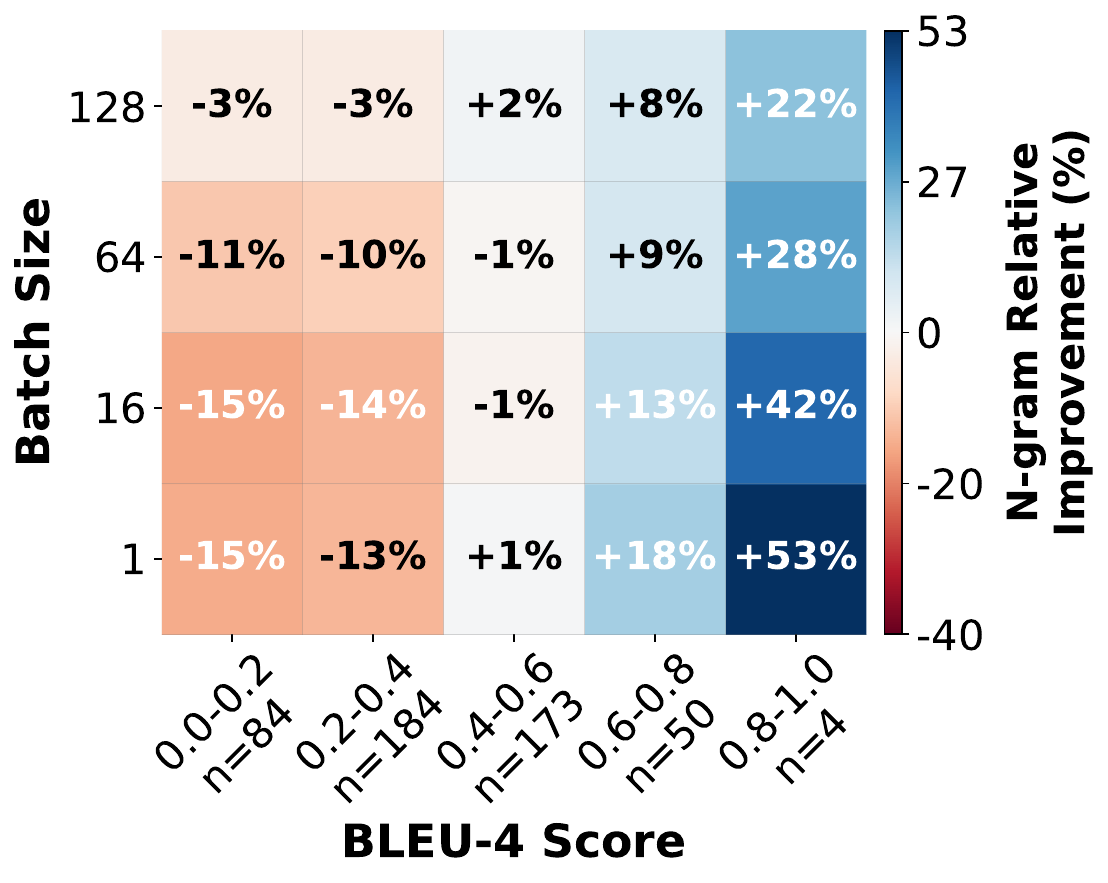}
\caption{Llama3.1-8B, $n$-gram-fixed-3 vs EAGLE}
\end{subfigure}
\begin{subfigure}{0.24\linewidth}
\includegraphics[width=\linewidth]{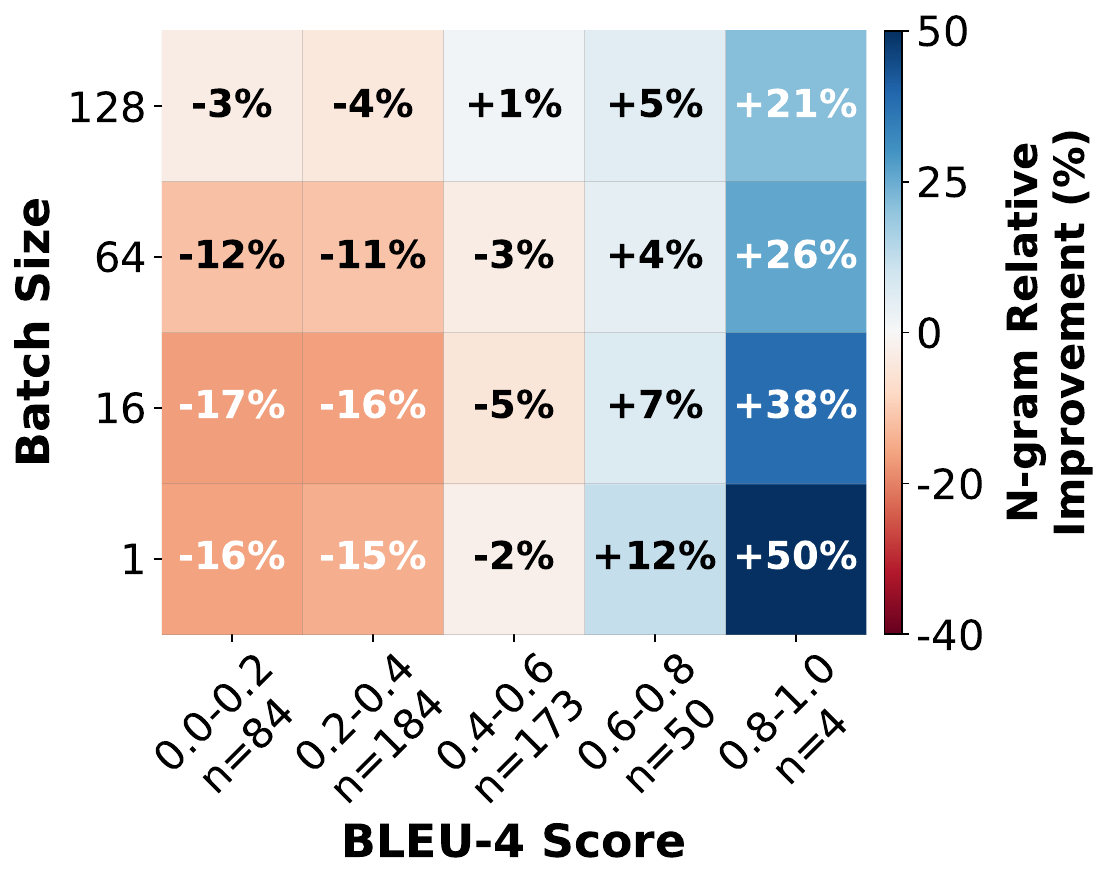}
\caption{Llama3.1-8B, $n$-gram-fixed-3 vs EAGLE-3}
\end{subfigure}
\begin{subfigure}{0.24\linewidth}
\includegraphics[width=\linewidth]{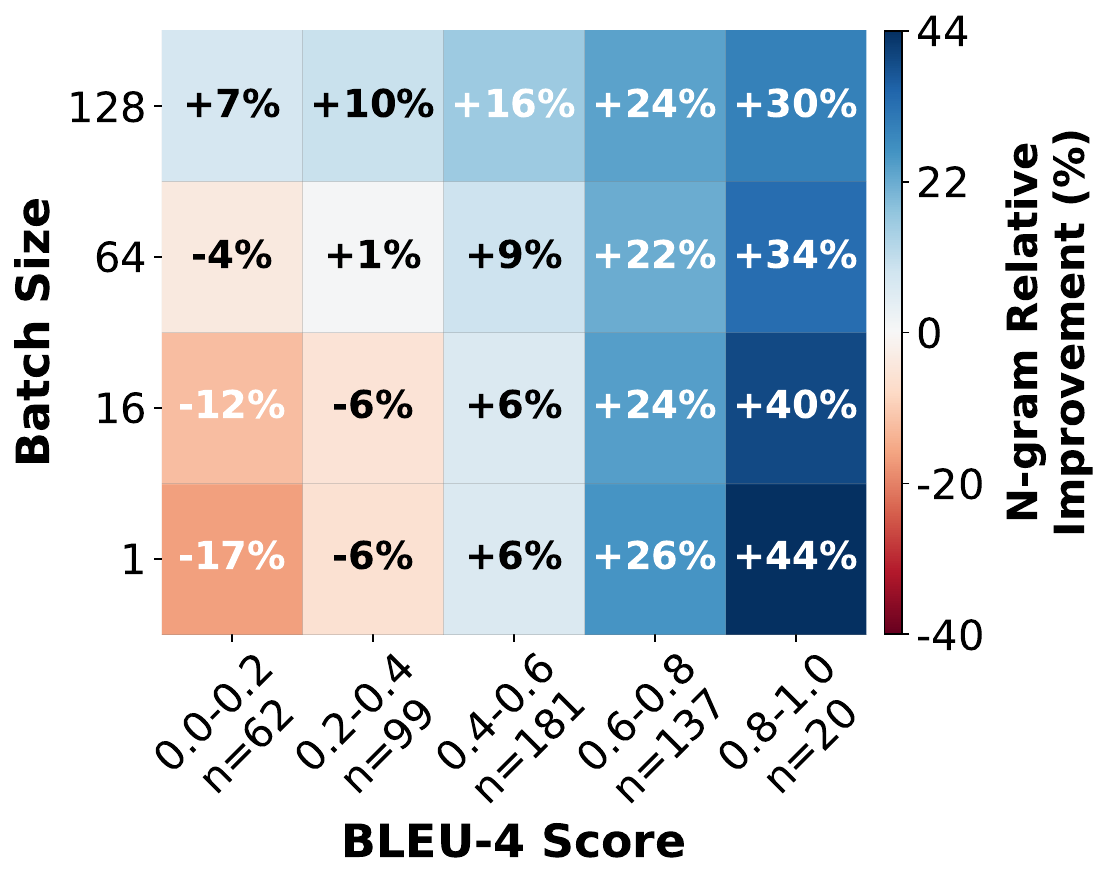}
\caption{Qwen3-8B, $n$-gram-fixed-3 vs EAGLE-3}
\end{subfigure}
\begin{subfigure}{0.24\linewidth}
\includegraphics[width=\linewidth]{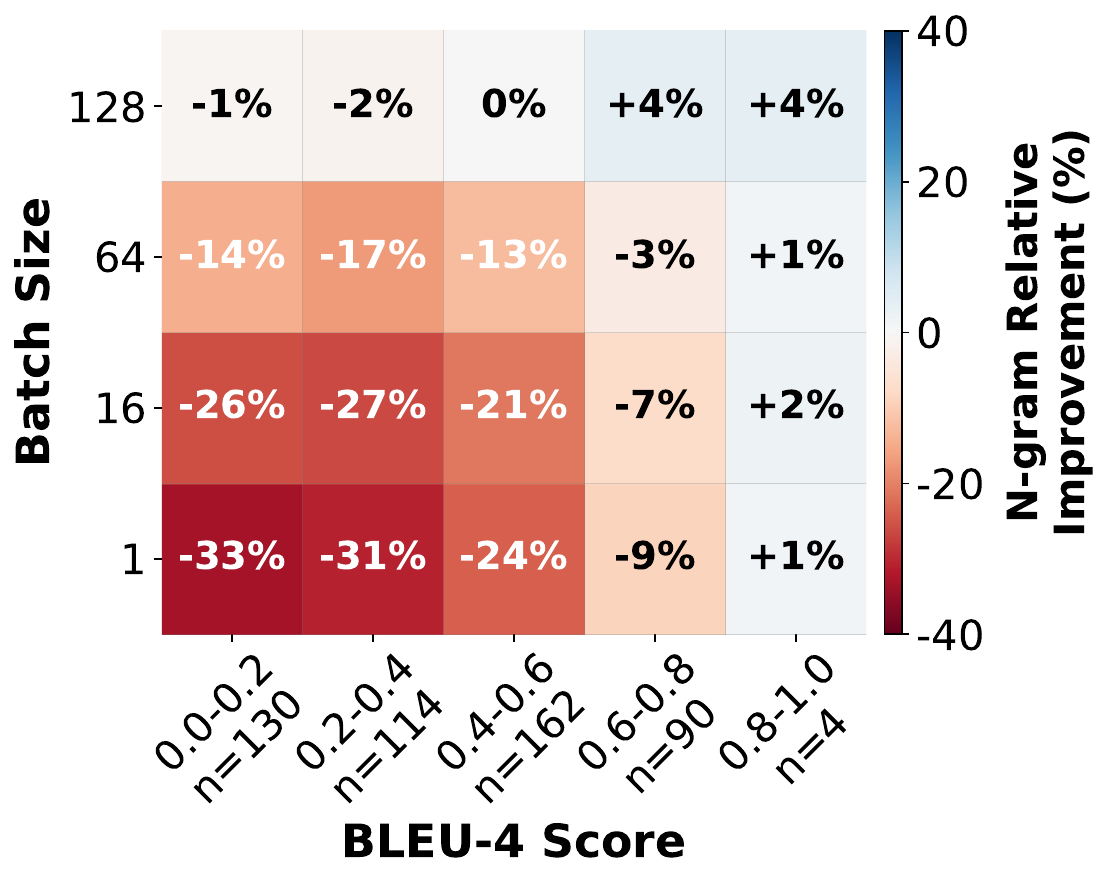}
\caption{Llama3-70B, $n$-gram-fixed-3 vs EAGLE}
\end{subfigure}

\begin{subfigure}{0.24\linewidth}
\includegraphics[width=\linewidth]{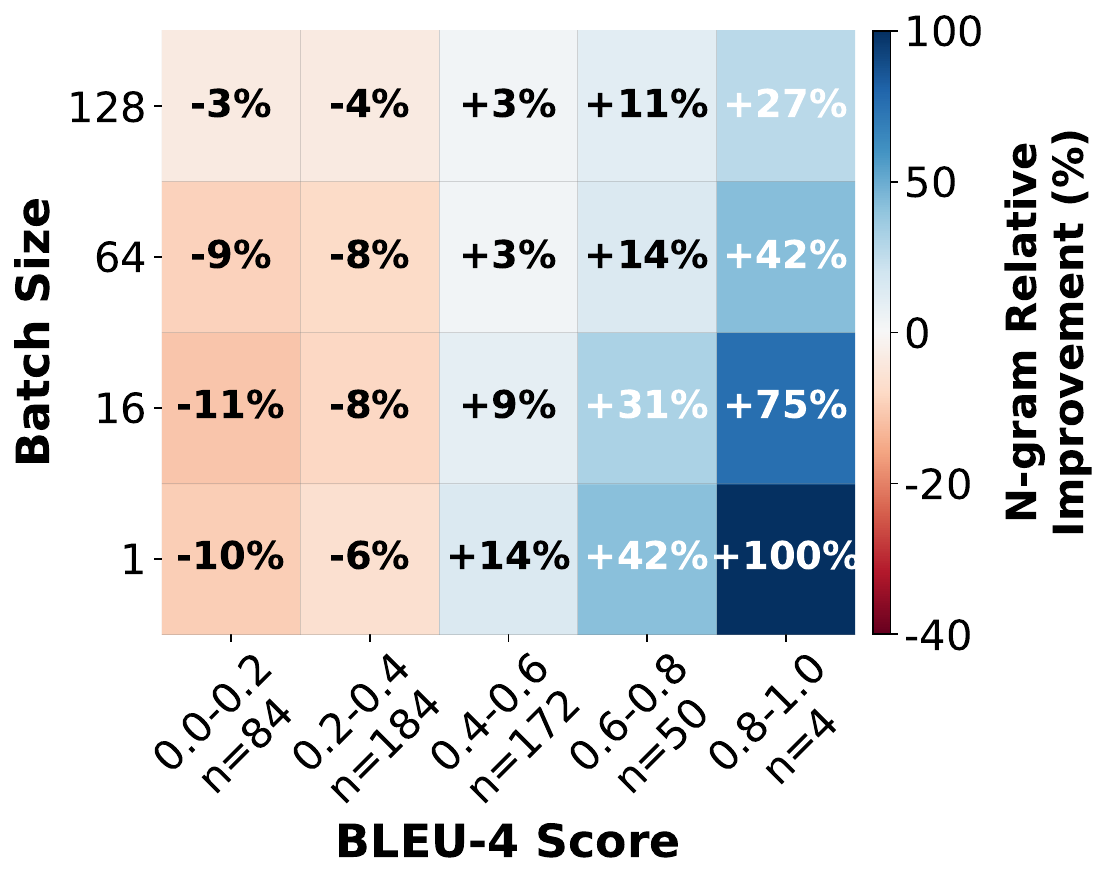}
\caption{Llama3.1-8B, $n$-gram-fixed-5 vs EAGLE}
\end{subfigure}
\begin{subfigure}{0.24\linewidth}
\includegraphics[width=\linewidth]{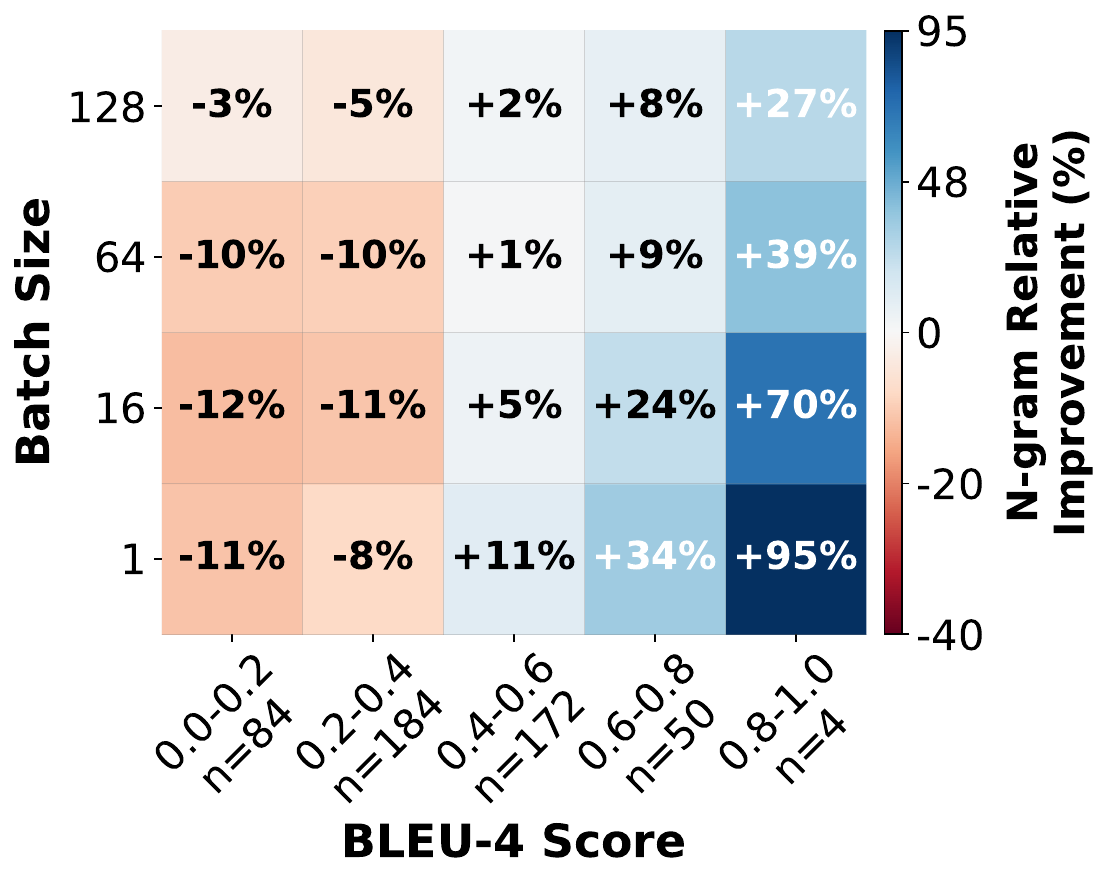}
\caption{Llama3.1-8B, $n$-gram-fixed-5 vs EAGLE-3}
\end{subfigure}
\begin{subfigure}{0.24\linewidth}
\includegraphics[width=\linewidth]{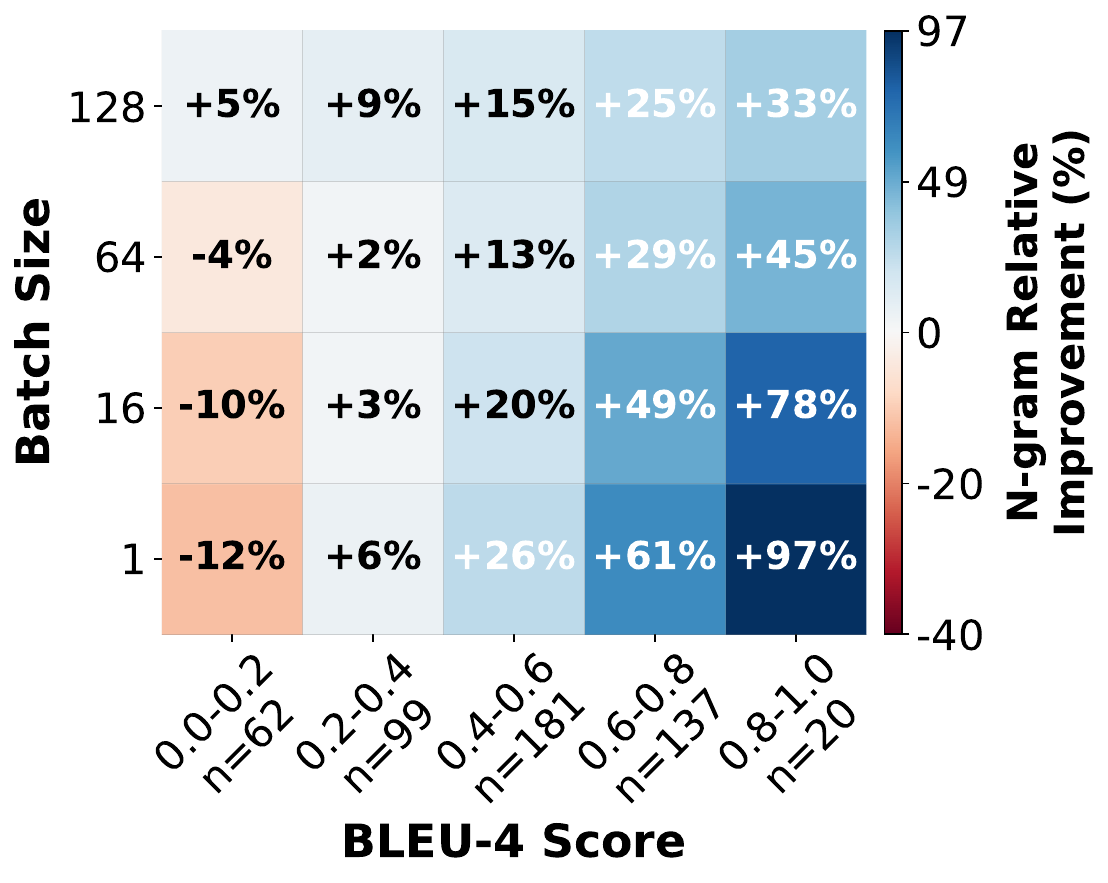}
\caption{Qwen3-8B, $n$-gram-fixed-5 vs EAGLE-3}
\end{subfigure}
\begin{subfigure}{0.24\linewidth}
\includegraphics[width=\linewidth]{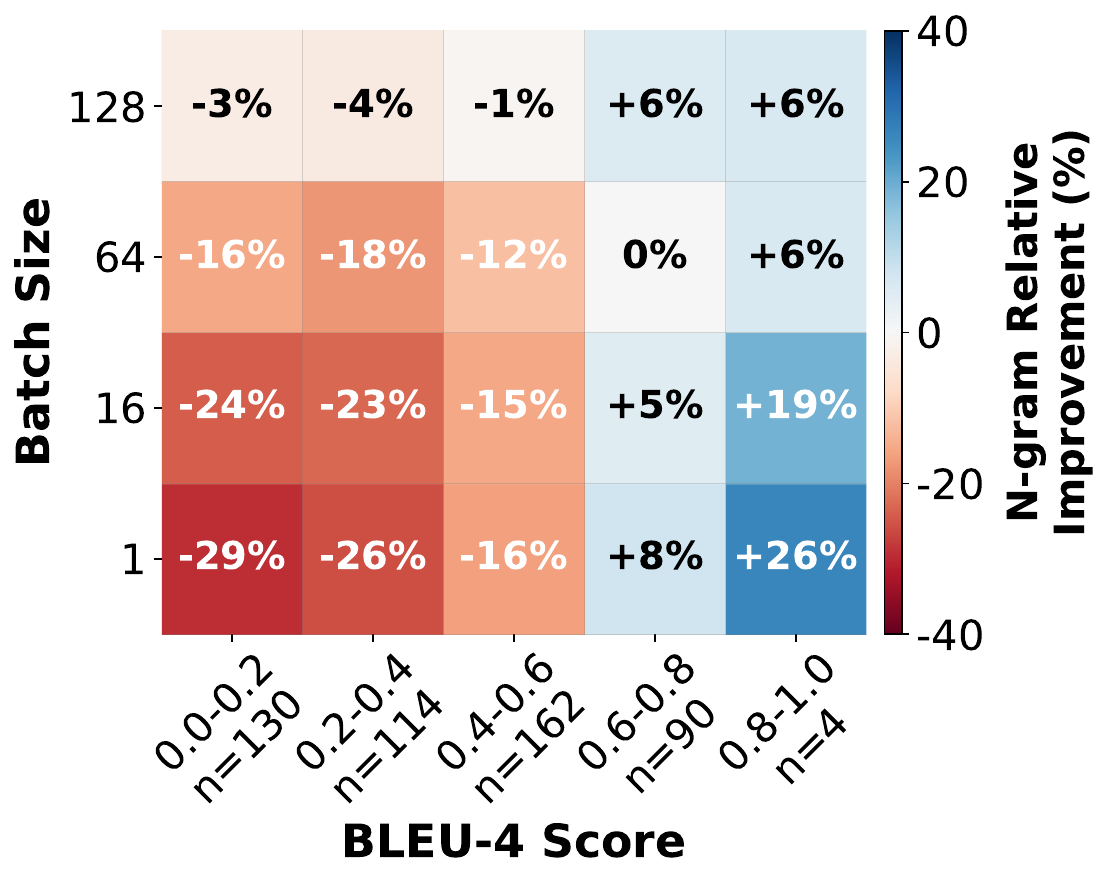}
\caption{Llama3-70B, $n$-gram-fixed-5 vs EAGLE}
\end{subfigure}

\caption{Correlation between BLEU-4 and $n$-gram speedup on InstructCoder. Each heatmap shows $n$-gram’s relative speedup (\%) over EAGLE or EAGLE-3, grouped by BLEU-4 scores (x-axis) and batch sizes (y-axis). Blue means that $n$-gram outperforms the draft-model method, while red means the opposite. $n{=}$XYZ in the x-axis labels denotes the number of requests in each BLEU bucket. Figures (a) to (d) show the results when $n$-gram proposes 3 tokens per step and Figures (e) to (h) show the results when that is set to 5.}
\label{fig:ngram_bleu_heatmap}
\vspace{-0.2in}
\end{center}
\end{figure*}

\section{Case Study: $n$-gram Performance on InstructCoder}
\label{subsect:ngram-bleu-score}

The $n$-gram SD is especially attractive for speculative decoding because it is training-free.
We observe that $n$-gram achieves particularly strong speedups on the code-editing workload, such as InstructCoder, and we therefore conduct a focused case study to understand the underlying reasons.

We hypothesize that its advantage stems from the \emph{local repetition} inherent in code-editing tasks: when upcoming tokens have already appeared in the prompt, $n$-gram lookup is more likely to propose correct continuations that are subsequently accepted by the target model during verification.
To test this hypothesis, we quantify prompt–output overlap using BLEU-$n$~\cite{papineni2002bleu}.
BLEU-$n$ measures the fraction of overlapping $n$-grams, with higher values indicating stronger local reuse or copying.
For each request, we compute the BLEU score between the prompt and the output generated without speculative decoding, and then group them into 5 intervals: requests with BLEU score of 0-0.2, 0.2-0.4, 0.4-0.6, 0.6-0.8 and 0.8-1.0.
Within each interval, we compare the speedups achieved by $n$-gram and by EAGLE/EAGLE-3.
We report the results using BLEU-4 score by default~\cite{papineni2002bleu}. In our experiments, BLEU-4 gives the clearest speedup heatmap and shows a clear cutoff where, above a certain overlap range, $n$-gram beats EAGLE/EAGLE-3 for every batch size. We have also repeated the analysis with BLEU-1 through BLEU-10 scores, and observe the same overall patterns.

\begin{quoting}[leftmargin=1.5em, rightmargin=1.5em, vskip=0pt]
\textbf{Overlap Drives $n$-gram Speedups.}
\emph{$n$-gram works best on code editing because the output often reuses short spans that already appear in the prompt. As overlap increases, $n$-gram speedup rises and eventually surpasses EAGLE/EAGLE-3.}
\end{quoting}
We observe that higher BLEU-$n$ scores strongly correlate with greater $n$-gram speedup (\autoref{fig:ngram_bleu_heatmap}). 
For requests with lower BLEU-$n$ scores (little overlap), $n$-gram underperforms due to inaccurate proposals. Once the BLEU-$n$ score exceeds a certain threshold, (e.g. larger than $0.6$ for Llama3.1-8B), $n$-gram consistently outperforms EAGLE and EAGLE-3 across all batch sizes. With a proposal length of 3, it achieves up to $53\%$ higher speedup than EAGLE and EAGLE-3. When we further increase the proposal length of $n$-gram to 5, the boundary where $n$-gram performs better is clearer, and this brings an even higher performance of up to $100\%$ higher speedup than EAGLE/EAGLE-3.
This is likely because a larger proposal length allows $n$-gram to reuse longer repeated spans in the prompt or recent context when the overlap is high. 
Overall, these results confirm that $n$-gram speculation benefits directly from prompt-level repetition in code-editing workloads. 

Finally, our BLEU analysis measures overlap only between the complete prompt and the complete generated output. In practice, $n$-gram can match and reuse tokens from the full in-context history, which includes both the prompt and the tokens generated so far. We leave an analysis that measures overlap with respect to this evolving context for future work.

\begin{figure*}[t]
  \centering
\begin{subfigure}[t]{0.35\linewidth}
\includegraphics[width=\linewidth]{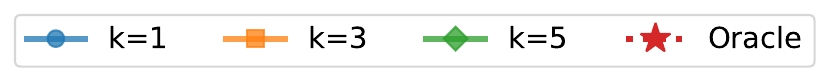}
\end{subfigure}

\begin{subfigure}[b]{0.22\linewidth}
    \includegraphics[width=\linewidth]{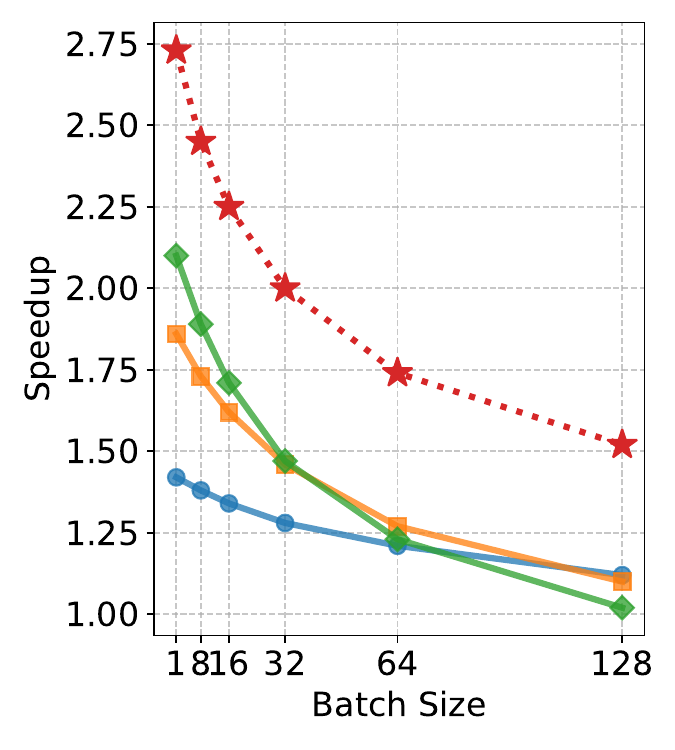}
    \caption{$n$-gram, InstructCoder.}
    \label{fig:speedup-oracle-ngram-instructcoder}
\end{subfigure}
\begin{subfigure}[b]{0.22\linewidth}
    \includegraphics[width=\linewidth]{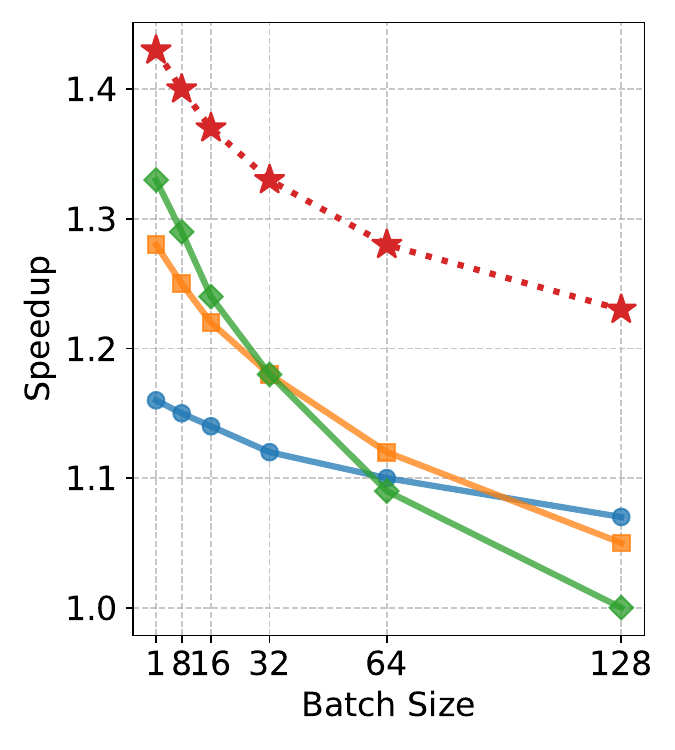}
    \caption{$n$-gram, ShareGPT.}
    \label{fig:speedup-oracle-ngram-sharegpt}
\end{subfigure}
\begin{subfigure}[b]{0.22\linewidth}
    \includegraphics[width=\linewidth]{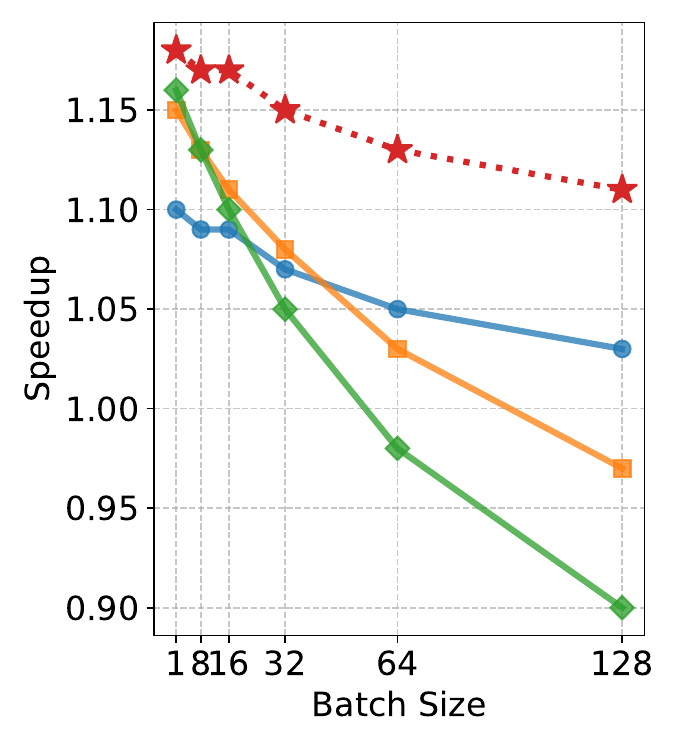}
    \caption{$n$-gram, GSM8K.}
    \label{fig:speedup-oracle-ngram-gsm8k}
\end{subfigure}
\begin{subfigure}[b]{0.22\linewidth}
    \includegraphics[width=\linewidth]{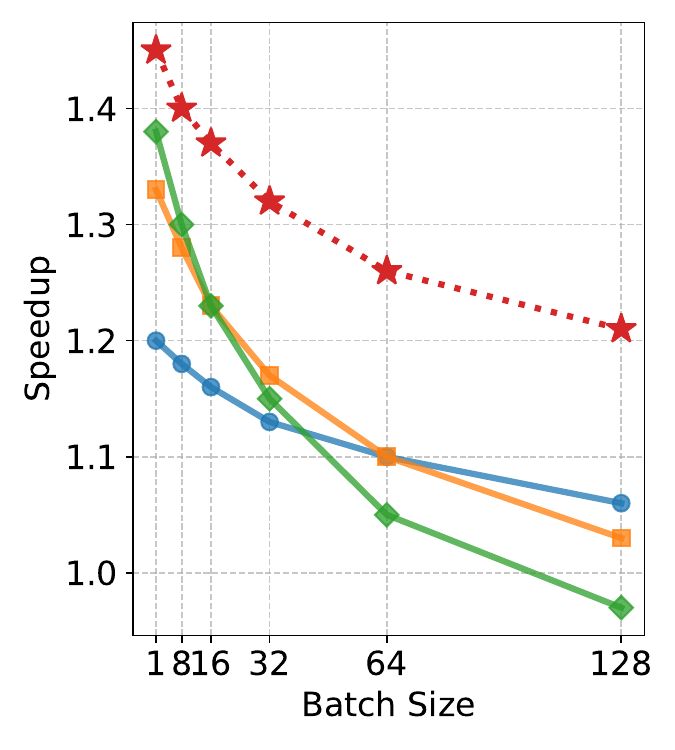}
    \caption{$n$-gram, CNN/DailyMail.}
    \label{fig:speedup-oracle-ngram-cnn}
\end{subfigure}

\begin{subfigure}[b]{0.22\linewidth}
    \includegraphics[width=\linewidth]{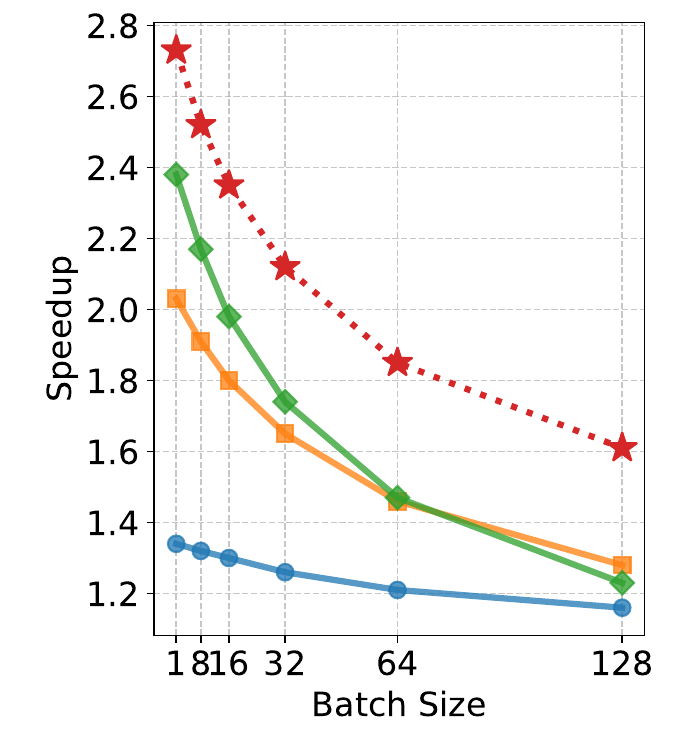}
    \caption{EAGLE, InstructCoder.}
    \label{fig:speedup-oracle-eagle-instructcoder}
\end{subfigure}
\begin{subfigure}[b]{0.22\linewidth}
    \includegraphics[width=\linewidth]{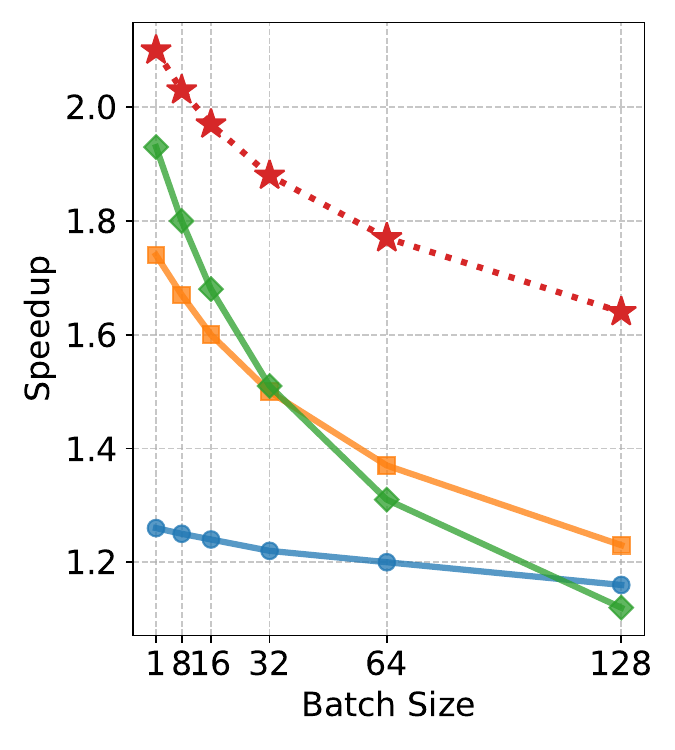}
    \caption{EAGLE, ShareGPT.}
    \label{fig:speedup-oracle-eagle-sharegpt}
\end{subfigure}
\begin{subfigure}[b]{0.22\linewidth}
    \includegraphics[width=\linewidth]{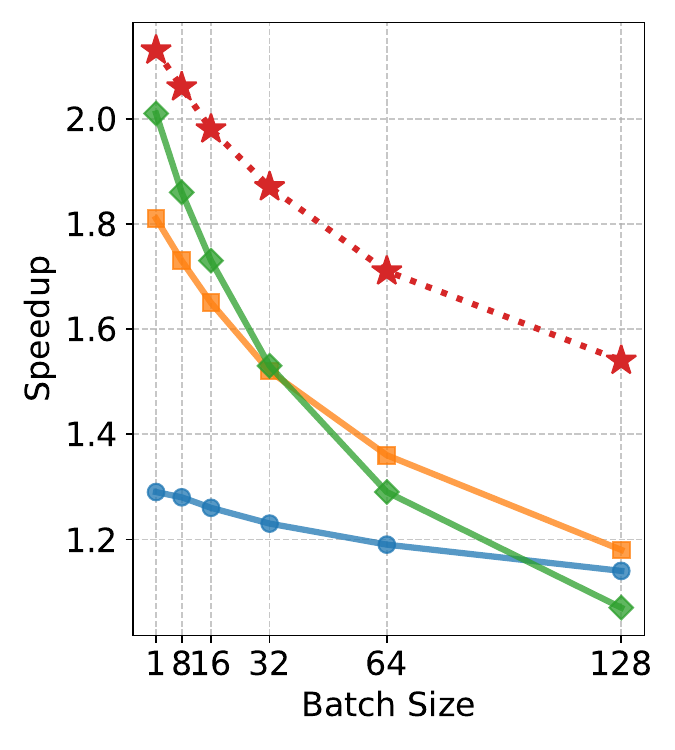}
    \caption{EAGLE, GSM8K.}
    \label{fig:speedup-oracle-eagle-gsm8k}
\end{subfigure}
\begin{subfigure}[b]{0.22\linewidth}
    \includegraphics[width=\linewidth]{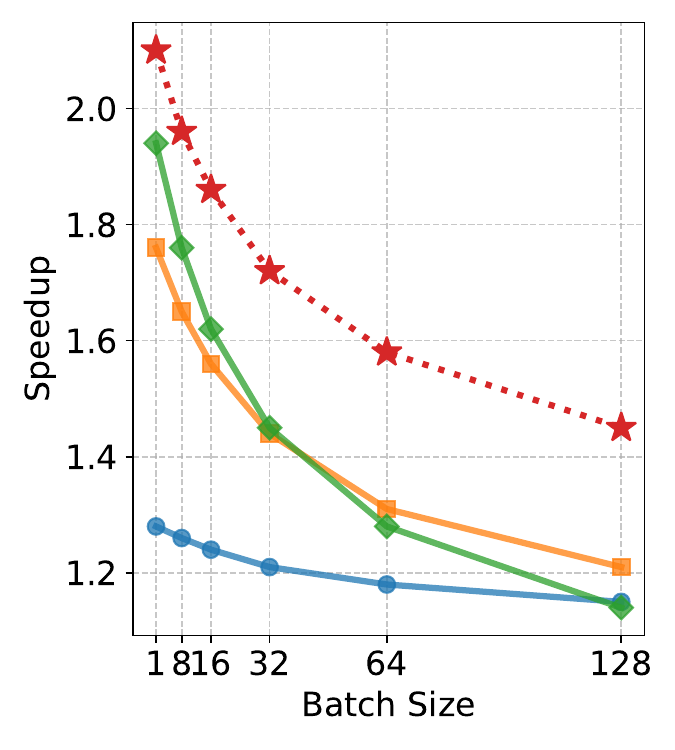}
    \caption{EAGLE, CNN/DailyMail.}
    \label{fig:speedup-oracle-eagle-cnn}
\end{subfigure}
    
    \caption{Oracle vs fixed proposed length speedup on Llama3.1-8B.}
    \label{fig:speedup-oracle}
%
\end{figure*}

\begin{figure*}[h]
  \centering
      \begin{subfigure}[b]{0.24\linewidth}
    \includegraphics[width=\linewidth]{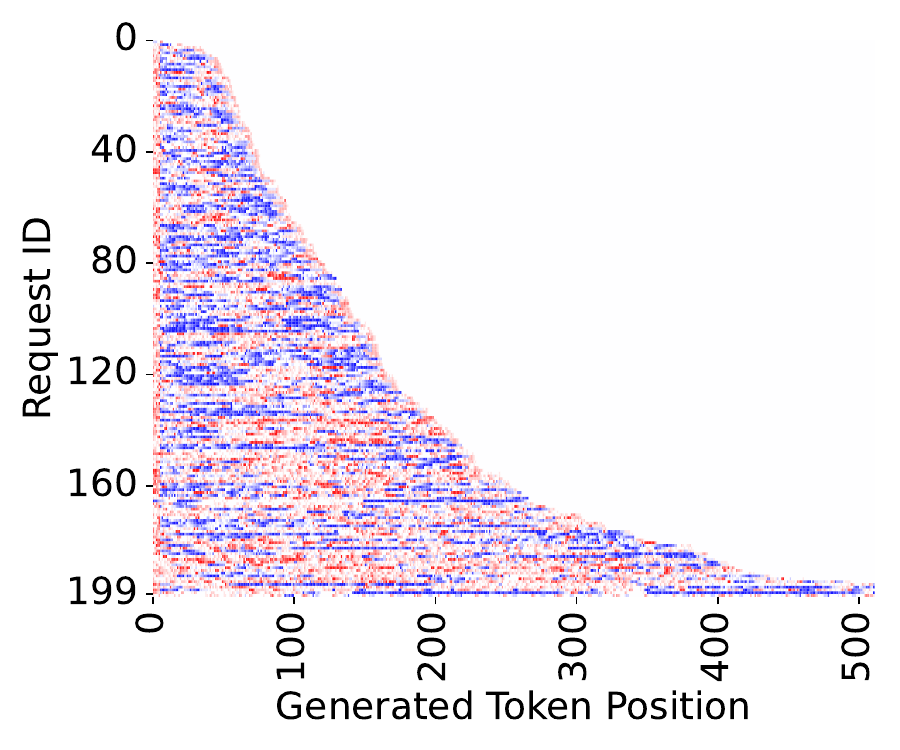}
    \caption{InstructCoder}
    \label{fig:acc-diff-instrcutcoder}
  \end{subfigure}
      \begin{subfigure}[b]{0.24\linewidth}
    \includegraphics[width=\linewidth]{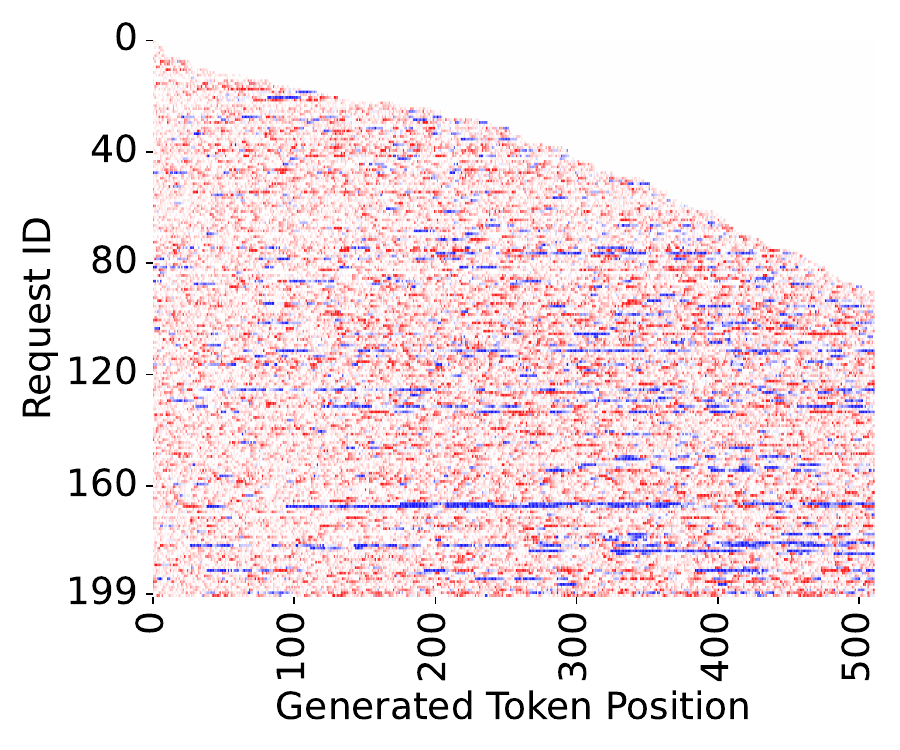}
    \caption{ShareGPT}
    \label{fig:acc-diff-sharegpt}
  \end{subfigure}
      \begin{subfigure}[b]{0.24\linewidth}
    \includegraphics[width=\linewidth]{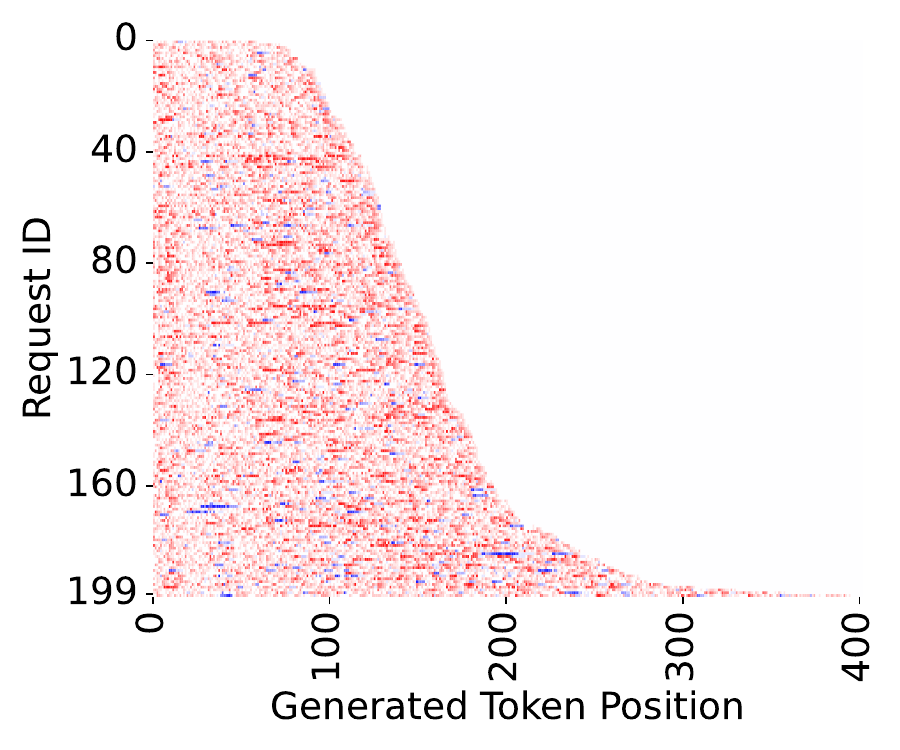}
    \caption{GSM8K}
    \label{fig:acc-diff-gsm8k}
  \end{subfigure}
          \begin{subfigure}[b]{0.24\linewidth}
    \includegraphics[width=\linewidth]{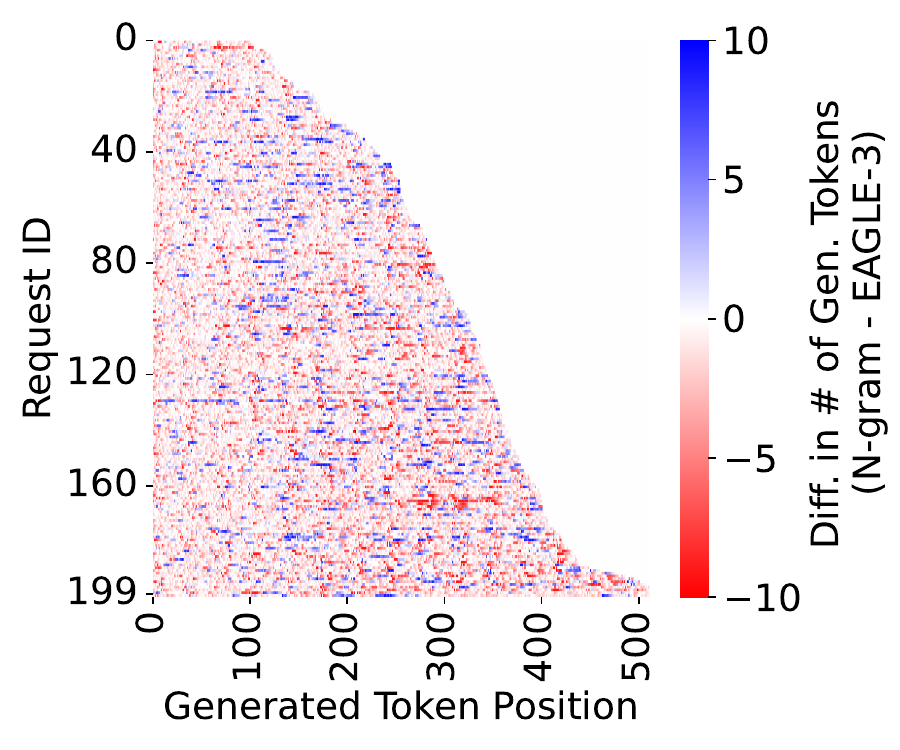}
    \caption{CNN/DailyMail}
    \label{fig:acc-diff-cnn}
  \end{subfigure}
  \caption{Per-position accepted-length difference between $n$-gram and EAGLE on Llama 3.1-8B. Red indicates positions where EAGLE accepts longer spans, while blue indicates positions favoring $n$-gram. The figure illustrates that different speculative decoding methods exhibit distinct acceptance behaviors across decoding positions.}
\label{fig:acc-diff}
\end{figure*}

\begin{figure*}
\centering
\begin{subfigure}[t]{0.65\linewidth}
\includegraphics[width=\linewidth]{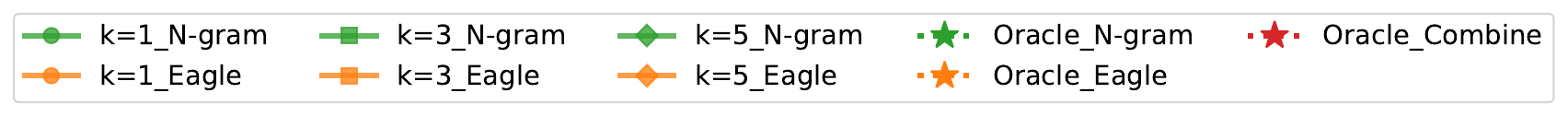}
\end{subfigure}

    \begin{subfigure}[b]{0.22\linewidth}
      \includegraphics[width=\linewidth]{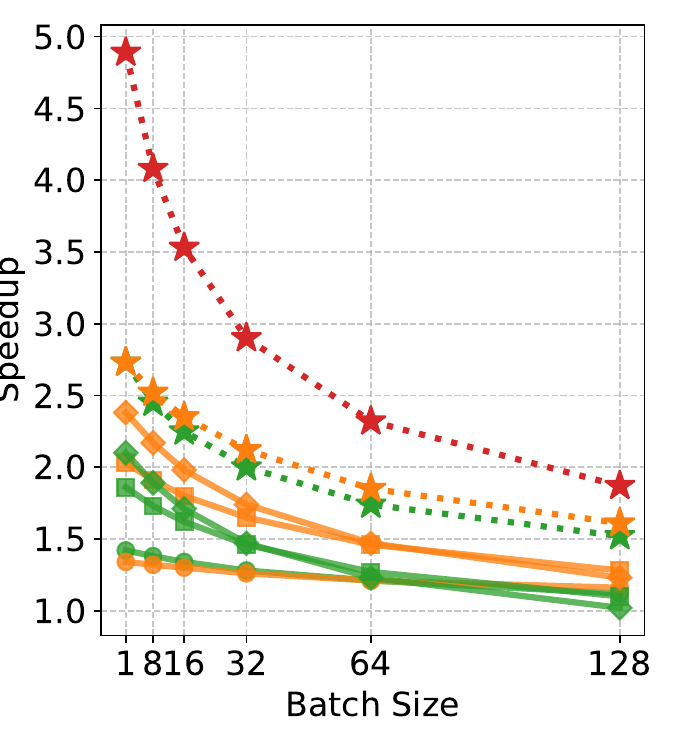}
      \caption{InstructCoder}
      \label{fig:opt-speedup-instrcutcoder}
  \end{subfigure}
  \begin{subfigure}[b]{0.22\linewidth}
      \includegraphics[width=\linewidth]{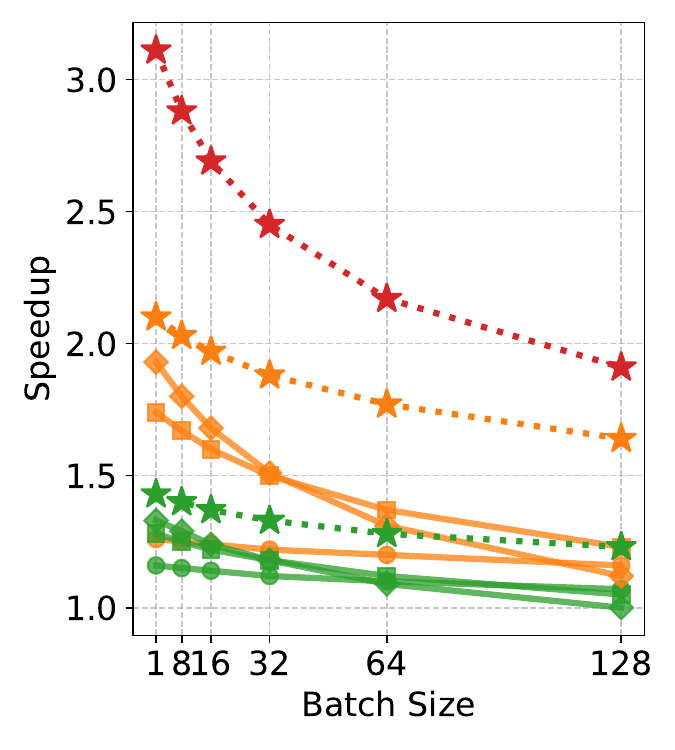}
      \caption{ShareGPT}
      \label{fig:opt-speedup-sharegpt}
  \end{subfigure}
   \begin{subfigure}[b]{0.22\linewidth}
      \includegraphics[width=\linewidth]{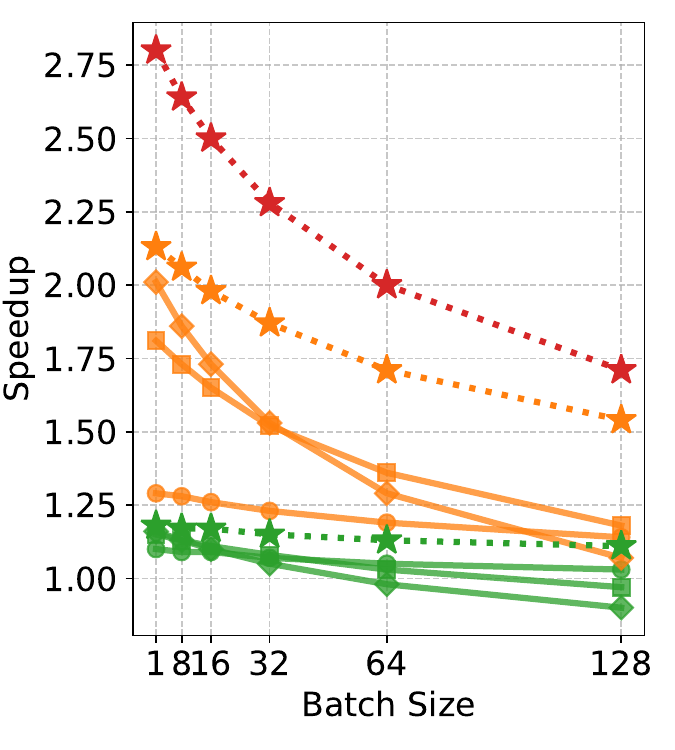}
      \caption{GSM8K}
      \label{fig:opt-speedup-gsm8k}
  \end{subfigure}
      \begin{subfigure}[b]{0.22\linewidth}
      \includegraphics[width=\linewidth]{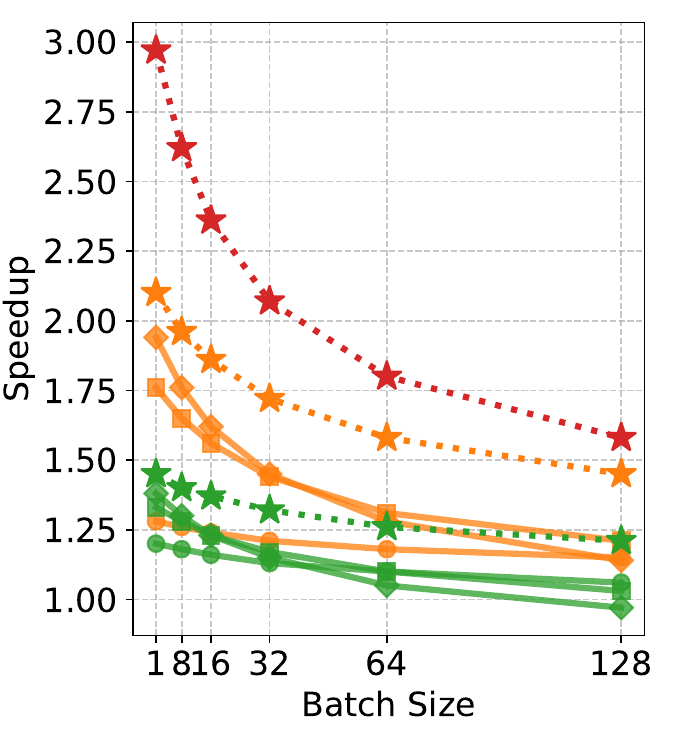}
      \caption{CNN/DailyMail}
      \label{fig:opt-speedup-cnn}
  \end{subfigure}
  \caption{Combining different SD methods for optimal speedup on Llama3.1-8B.}
  \label{fig:opt-speedup}
\end{figure*}

\section{Theoretical Speedup Upper Bound of Speculative Decoding}
\label{sec:theoretical-speedup}
In this section, we aim to derive an upper bound of the speedup achievable by SD. Our goal is to understand how far current SD methods are from the theoretical optimum, assess their current performance limits, and highlight potential directions for future research.

As discussed in \autoref{sec:breakdown}, verification time dominates the overall execution cost. This naturally raises the question: \emph{if we could eliminate wasted verification by avoiding the verification of tokens that would ultimately be rejected, what is the maximum speedup that speculative decoding could achieve?
In other words, assuming an oracle—or a sufficiently accurate mechanism—that proposes only tokens likely to be accepted, what is the upper bound on the achievable speedup?} In the following section, we try to derive this number and understand the gap between current speedup and theoritical upper bound.

Moreover, as observed in \autoref{sec:acc-behavior}, different SD variants exhibit distinct acceptance behaviors. Specifically, they differ in the number of tokens accepted at each generation step. To better understand the true performance upper bound, we investigate whether combining different SD strategies could yield an even more optimal acceptance pattern and, consequently, higher overall efficiency. 


\subsection{Minimal Verification Cost}
We begin by addressing the following question: if we could minimize the verification cost, what is the maximum speedup that speculative decoding could achieve? To explore this, we compare the performance of using a fixed proposed length against that of an oracle-based proposed length. The oracle setup assumes that the number of tokens accepted at each generation step is known in advance. At each step, we set the proposed length equal to the actual accepted length, ensuring that all proposed tokens are accepted. This setup effectively simulates the ideal case where verification incurs no rejection overhead.

One might argue that if all proposed tokens can be perfectly accepted, the large model would no longer be necessary. While that is theoretically correct, it is practically unattainable—no proposal mechanism can perfectly predict future target tokens. Hence, the oracle configuration serves purely as an upper bound on the achievable speedup of speculative decoding, highlighting the performance ceiling that real methods can approximate.

We conduct this analysis using Llama 3.1-8B on the InstructorCoder, ShareGPT, GSM8K, and CNN/DailyMail datasets. We observe a substantial gap between the oracle speedup and the speedup achieved with a fixed proposal length. The oracle speedup represents an upper bound in which the accepted length at each decoding step is known in advance, and the system proposes exactly that many tokens. As a result, every proposed token is accepted, eliminating any wasted speculation.

 As shown in \autoref{fig:speedup-oracle}, on the Instructcoder dataset with a batch size of one and $n$-gram as the SD method, the oracle setup achieves a speedup of approximately 2.75$\times$, whereas the fixed proposed length configuration attains about 2.1$\times$ for the best proposed length of 5. Moreover, the gap generally widens as batch size increases: the fixed-$k$ curves drop much faster, while the oracle speedup degrades more gently. This trend is consistent with our observation in \autoref{fig:exec-breakdown}. As batch size grows, verification takes up a larger share of the decoding time, so fixed-$k$ methods would incur larger overhead for verifying draft tokens that are ultimately rejected.

\subsection{An Adaptive Approach to Achieve Optimal Speedup}
Although EAGLE generally outperforms the $n$-gram method, 
a closer inspection of acceptance behavior reveals complementary strengths between the two. 
As illustrated in \autoref{fig:acc-diff}, the per-position differences in accepted tokens exhibit alternating regions of red (favoring EAGLE) and blue (favoring $n$-gram), indicating that the two methods excel under different conditions. This behavior is intuitive: for example, in code snippets with strong local regularities or recurring grammatical patterns, $n$-gram can often find exact matches, which are likely to be correct.
In such cases, because $n$-gram incurs substantially lower proposing overhead, it can be the more efficient choice. At other positions where local repetition is weaker, EAGLE serves as a robust fallback, since it proposes correct tokens across diverse contexts more consistently.

This naturally brings up the next question: \emph{can we combine their advantages to achieve even higher overall speedups?}

To explore the upper bound on achievable speedup, we assume a \emph{perfect} predictor that can (1) determine which method performs best at each position, and (2) accurately predict the number of tokens that will be accepted at that position. Under this idealized assumption, we measure the maximum possible speedup attainable by speculative decoding.

It is important to note the simplifications in this setup, which represent an upper bound on achievable performance. First, we assume the predictor is flawless in determining the better method for each position. Second, we presume perfect knowledge of the number of tokens each method can accept at every position. 
Finally, we do not account for the overhead of maintaining the KV cache for EAGLE's proposing head, which would otherwise require partial prefills if a request was paused and resumed.

Under these assumptions, we report the best achievable speedup in \autoref{fig:opt-speedup}. \textbf{Oracle\_Combine} represents a theoretical upper bound in which, at each generation position, the method (EAGLE or $n$-gram) that yields the longer accepted span is selected. The resulting accepted length is then used as the proposed length. Compared to the best fixed strategy, \textbf{Oracle\_Combine} exposes substantial additional headroom, achieving up to a $2.2\times$ further speedup.

The size of this headroom depends strongly on the workload. InstructCoder shows the largest gap between existing methods and Oracle\_Combine, matching \autoref{fig:acc-diff} where red and blue regions alternate frequently, i.e., there are many positions where one method clearly outperforms the other. ShareGPT and CNN/DailyMail exhibit similar but smaller gains, indicating more limited but still meaningful room for the methods to complement each other. In contrast, GSM8K offers little additional benefit from combining because $n$-gram rarely achieves long acceptances, so \textbf{Oracle\_Combine} stays close to \textbf{Oracle\_Eagle}. 

Taken together, these results point to a promising direction for future work: developing an accurate yet lightweight predictor capable of adapting to varying levels of acceptance behavior across workloads, requests and token positions.



\section{Conclusion}
This work presents the first systematic evaluation of speculative decoding (SD) in a real, optimized inference engine. By systematically dissecting end-to-end performance across workloads and SD variants, we identify verification as the dominant bottleneck and reveal strong variability in acceptance behavior across positions, requests, and datasets. Leveraging these insights, we quantify the theoretical upper bound of SD speedup and expose the gap between observed and ideal efficiency. Together, these findings deepen our understanding of SD’s practical behavior and highlight opportunities that can further unlock its full potential in large-scale inference systems.

\section{Acknowledgements}
We are grateful to Tomas Ruiz for implementing draft-model-based speculative decoding in vLLM, which allowed us to profile and compare this approach. We also thank Yilong Zhao, Yifan Qiao, and other members of the Sky Computing Lab for their helpful discussions and feedback. This work was supported in part by a gift from NVIDIA, including the DGX server used in this study. Jiaxiang is supported by the Singapore National Science Scholarship and the NUS Development Grant. Any opinions, findings, and conclusions or recommendations expressed in this paper are those of the authors and do not necessarily reflect the views of the supporting organizations.
\bibliography{example_paper}
\bibliographystyle{mlsys2025}

\clearpage
\appendix
\section{Appendix}

\subsection{Dataset}
\label{appendix:dataset}
\begin{figure}[h]
    \centering
    \includegraphics[width=\linewidth]{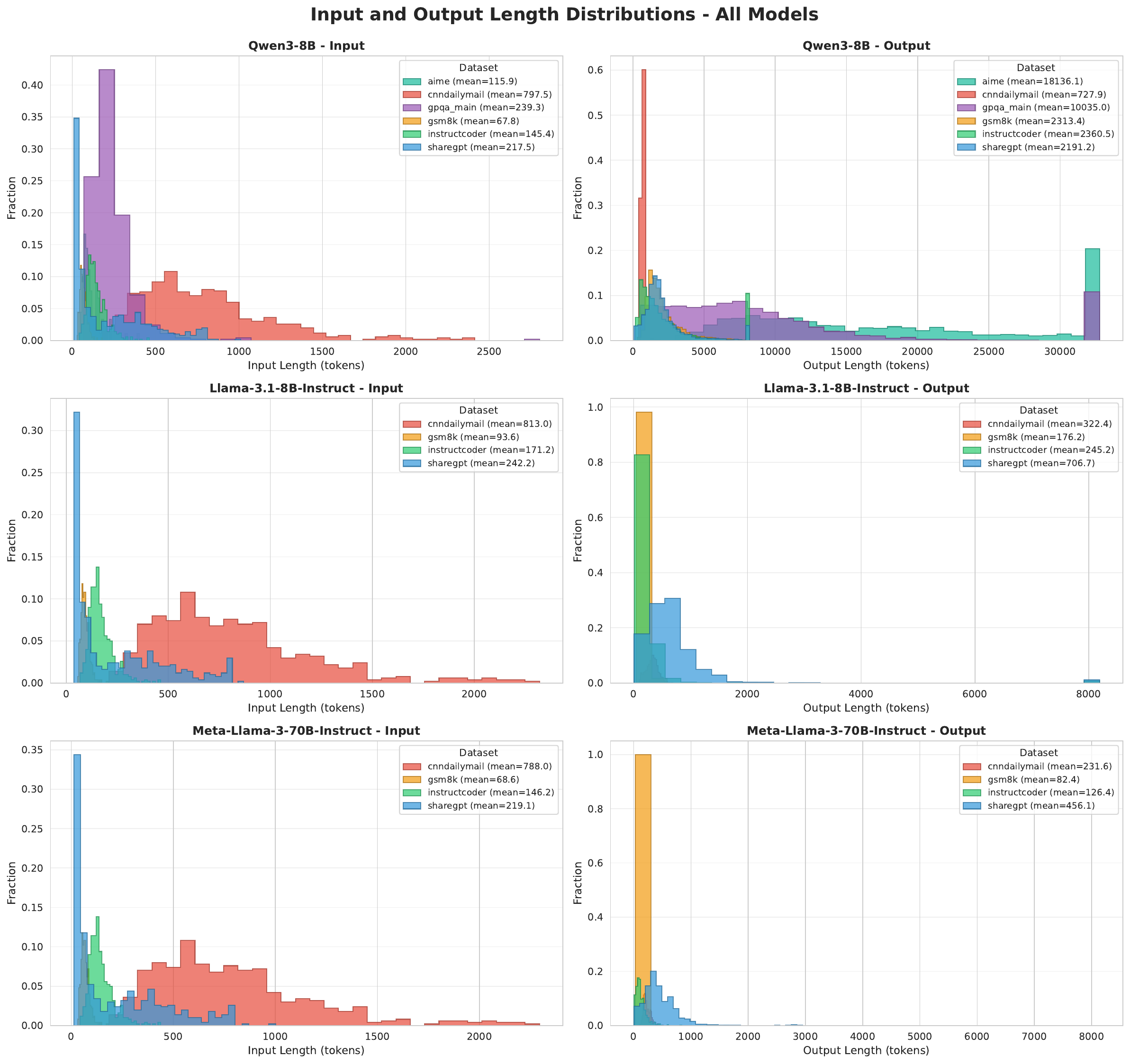}
    \caption{Dataset Input/Output Length distributions.}
    \label{fig:dataset-input-output-plot}
\end{figure}

\subsection{Memory Breakdown Calculations}
\label{appendix:mem}
\paragraph{Static memory (GiB).} Assuming FP16 weights (2 bytes/param), the static GPU memory for model weights:
\[
M_\text{static,GiB}=\frac{(P_\text{target}+P_\text{draft})\cdot 2}{2^{30}}
\]
Here, $P_\text{target}$ is the target model's parameter size and $P_\text{draft}$ includes any additional parameters introduced by the speculative decoding method, both in units of Billions.

For EAGLE/EAGLE3, $P_\text{draft}$ refers to the parameters used in speculative decoding. For the EAGLE models we used, $P_\text{draft}$ refers to the parameter size of the autoregressive head, which includes one decoding layer and one fully connected (FC) layer. They share the same embedding layer and language modeling (LM) head with the target model. For the EAGLE3 models we used, $P_\text{draft}$ refers to the parameter size of the entire EAGLE3 model. It includes one decoding layer, the multi-layer feature fusion FC layer, a final normalization layer and its own LM head. In addition, the model checkpoints include two 1-dimensional token-id remapping tables (\texttt{t2d} and \texttt{d2t}) that translate between the target model’s vocabulary and the draft LM head’s vocabulary; they typically contribute negligibly to the parameter memory. When loading model weights in vLLM v0.10.1.1, the \texttt{t2d} tensors are skipped, and we exclude them when counting the parameter size. Similarly, these EAGLE3 models reuse the same embedding layer from the target models.

For Qwen3-0.6B, \textit{tie\_word\_embeddings} defaults to True and its LM head shares the same weight matrix as the embedding layer; thus, we exclude the size of its language modeling (LM) head.

For example, the static memory for Llama3-70B-Instruct with Llama3.2-1B-Instruct as the draft model is $(70.55+1.23)*10^9*2/2^{30}=~133.7GiB$. For Llama3.1-8B-Instruct with EAGLE-LLaMA3.1-Instruct-8B, it is $(8.03+0.25)*10^9*2/2^{30}=15.42GiB$.

\paragraph{Per-token KV cache (KiB).}
To calculate the KV cache size of each generated token:
\[
M_\text{KV/token,KiB}=
\frac{L_\text{h}\cdot 2 \cdot n_\text{kv}\cdot d_\text{head}\cdot 2}{2^{10}} .
\]
$L_\text{h}$ is the number of hidden layers,
$n_\text{kv}$ is the number of KV heads (since our models are all Grouped-Query-Attention-based),
$d_\text{head}$ is the head dimension, the first factor $2$ is for both Key and Value, and the last factor $2$ is bytes/element for FP16. For Llama3-70B-Instruct without speculative decoding or model-free SD methods like $n$-gram, it is $80*2*8*128*2/2^{10}=~320KiB$. Similarly, for Llama3.1-8B-Instruct, it is $32*2*8*128*2/2^{10}=~128KiB$.

For draft-model-based SD, the total per-token KV cache is the sum over each generated token by the target and draft model. For Llama3-70B-Instruct with Llama3.2-1B-Instruct as the draft model, the per-token KV cache memory size is $80*2*8*128*2/2^{10}$ + $16*2*8*64*2/2^{10}$ = $352KiB$.

Similarly, for EAGLE-based SD, the total per-token KV cache is the sum over each generated token by the target and EAGLE heads. For Llama3-70B-Instruct with Llama3.2-1B-Instruct as the draft model, the per-token KV cache memory size is $80*2*8*128*2/2^{10}$ + $1*2*8*128*2/2^{10}$ = $324KiB$. The value 1 here means that EAGLE uses an additional transformer block for proposal generation, which effectively adds one extra layer.

\begin{table}[t]
\centering
\small
\setlength{\tabcolsep}{4pt}
\begin{tabular}{l|rrrr}
\toprule
Model & $P$ & $L$ & $n_\text{kv}$ & $d_\text{head}$ \\
\midrule
Llama3.1-8B-Instruct  & 8.03B & 32 & 8 & 128 \\
Llama3-70B-Instruct   & 70.55B & 80 & 8 & 128 \\
Llama3.2-1B-Instruct  & 1.23B & 16 & 8 & 64 \\
Qwen3-8B              & 8.19B & 36 & 8 & 128 \\
Qwen3-0.6B              & 0.596B & 28 & 8 & 128 \\
EAGLE-LLaMA3.1-Instruct-8B & 0.25B & 1 & 8 & 128 \\
EAGLE3-LLaMA3.1-Instruct-8B & 0.425B & 1 & 8 & 128 \\
EAGLE-LLaMA3-Instruct-70B & 0.99B & 1 & 8 & 128 \\
EAGLE3-Qwen3-8B            &  0.40B     & 1 & 8 & 128 \\
\bottomrule
\end{tabular}
\vspace{2pt}
\caption{Model specs used for memory calculations. $P$ is the parameter size of the weights used, obtained by counting their parameters; $L$ is the number of hidden layers, $n_\text{kv}$ is the number of Key/Value heads, and $d_\text{head}$ is the head dimension, obtained from the model's configuration files on Hugging Face.}
\label{tab:mem_model_specs}
\vspace{-0.10in}
\end{table}

\subsection{Acceptance Behaviour}
\label{appendix:acc-behaviour}



\begin{figure}[h]
  \centering
  \includegraphics[width=0.7\linewidth]{figures/acc_len/request_level/legend.pdf}
  
    \begin{subfigure}{0.3\linewidth}
    \includegraphics[width=\linewidth]{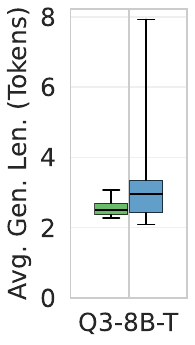}
    \caption{AIME22-24}
  \end{subfigure}
      \hspace{5pt}
    \begin{subfigure}{0.3\linewidth}
    \includegraphics[width=\linewidth]{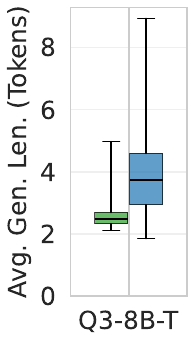}
    \caption{GPQA-Main}
  \end{subfigure}
\caption{Request-level generated length across datasets and models. Each box shows the distribution of per-request mean accepted length, with the box spanning the 25th–75th percentiles and whiskers covering the 5th–95th percentiles. Model used is Qwen3-8B-Thinking.}
  \label{fig:acc-len-request-level-box-plot-reasoning}
\end{figure}

\begin{table}[h!]
\centering
\begin{tabular}{llccc}
\toprule
Dataset & Method & Mean & Median & Std \\
\midrule
InstructCoder & $n$-gram & 7.27 & 6.57 & 4.19 \\
InstructCoder & EAGLE & 4.24 & 4.21 & 1.07 \\
InstructCoder & EAGLE-3 & 4.18 & 4.06 & 0.99 \\
CNN/DailyMail & EAGLE-3 & 3.08 & 2.97 & 0.61 \\
CNN/DailyMail & EAGLE & 2.32 & 2.31 & 0.21 \\
CNN/DailyMail & $n$-gram & 2.33 & 1.96 & 1.08 \\
ShareGPT & EAGLE-3 & 3.06 & 2.92 & 0.87 \\
ShareGPT & EAGLE & 2.82 & 2.68 & 0.80 \\
ShareGPT & $n$-gram & 2.02 & 1.37 & 1.87 \\
GSM8K & EAGLE-3 & 3.02 & 2.92 & 0.47 \\
GSM8K & EAGLE & 2.56 & 2.51 & 0.25 \\
GSM8K & $n$-gram & 1.41 & 1.28 & 0.45 \\
\midrule
GPQA\_Main & $n$-gram & 4.20 & 3.73 & 2.24 \\
GPQA\_Main & EAGLE-3 & 2.78 & 2.49 & 0.87 \\
AIME22-24 & $n$-gram & 3.53 & 2.96 & 2.29 \\
AIME22-24 & EAGLE-3 & 2.62 & 2.50 & 0.54 \\
\bottomrule
\end{tabular}
\caption{Dataset-level generated length by method, aggregated across models. For non-reasoning workloads, the average is aggregated over the results for Llama3.1-8B and Llama3-70B. For reasoning workloads, the average is aggregated over that of Qwen3-8B-Thinking.}
\label{tab:dataset_agg_by_method}
\end{table}

\clearpage


\begin{table*}[t!]
\centering
\caption{Main experiment configurations across datasets, models, and speculative decoding (SD) methods used in End-to-End Speedup Measurement. DM stands for Draft-Model-Based SD.}
\label{tab:exp-config}
\setlength{\tabcolsep}{2.5pt}
\renewcommand{\arraystretch}{1.1}
\begin{tabular}{l l l l c c}
\toprule
\textbf{Dataset} & \textbf{Model} & \textbf{Hardware} & \textbf{SD Method} & \textbf{Max Output Len.} & \textbf{Batch Size}\\
\midrule
\texttt{CNNDailyMail} & Llama3.1-8B-Instruct & 1$\times$H100 & $n$-gram, EAGLE, EAGLE-3  & 8K & 1–128 \\
\texttt{ShareGPT}     & Llama3.1-8B-Instruct & 1$\times$H100 & $n$-gram, EAGLE, EAGLE-3  & 8K & 1–128 \\
\texttt{InstructCoder}& Llama3.1-8B-Instruct & 1$\times$H100 & $n$-gram, EAGLE, EAGLE-3  & 8K & 1–128 \\
\texttt{GSM8K}        & Llama3.1-8B-Instruct & 1$\times$H100 & $n$-gram, EAGLE, EAGLE-3  & 8K & 1–128 \\
\midrule
\texttt{CNNDailyMail} & Llama3-70B-Instruct  & 4$\times$H100 & $n$-gram, EAGLE, DM       & 8K & 1–128 \\
\texttt{ShareGPT}     & Llama3-70B-Instruct  & 4$\times$H100 & $n$-gram, EAGLE, DM         & 8K & 1–128 \\
\texttt{InstructCoder}& Llama3-70B-Instruct  & 4$\times$H100 & $n$-gram, EAGLE, DM         & 8K & 1–128 \\
\texttt{GSM8K}        & Llama3-70B-Instruct  & 4$\times$H100 & $n$-gram, EAGLE, DM         & 8K & 1–128 \\
\midrule
\texttt{CNNDailyMail} & Qwen3-8B    & 1$\times$H100 & $n$-gram, EAGLE-3, DM (v0.11.1rc1)         & 8K  & 1–128 \\
\texttt{ShareGPT}     & Qwen3-8B    & 1$\times$H100 & $n$-gram, EAGLE-3, DM (v0.11.1rc1)       & 8K  & 1–128 \\
\texttt{InstructCoder}& Qwen3-8B    & 1$\times$H100 & $n$-gram, EAGLE-3, DM (v0.11.1rc1)     & 8K  & 1–128 \\
\texttt{GSM8K}        & Qwen3-8B    & 1$\times$H100 & $n$-gram, EAGLE-3, DM (v0.11.1rc1)     & 8K  & 1–128 \\
\midrule
\texttt{CNNDailyMail} & Qwen3-8B-Thinking    & 1$\times$H100 & $n$-gram, EAGLE-3         & 8K  & 1–128 \\
\texttt{ShareGPT}     & Qwen3-8B-Thinking    & 1$\times$H100 & $n$-gram, EAGLE-3      & 8K  & 1–128 \\
\texttt{InstructCoder}& Qwen3-8B-Thinking    & 1$\times$H100 & $n$-gram, EAGLE-3    & 8K  & 1–128 \\
\texttt{GSM8K}        & Qwen3-8B-Thinking    & 1$\times$H100 & $n$-gram, EAGLE-3    & 8K  & 1–128 \\
\midrule
\multicolumn{6}{l}{\textit{Reasoning workloads}} \\
\texttt{AIME(22--24)} & Qwen3-8B-Thinking    & 1$\times$H100 & $n$-gram, EAGLE-3       & 32K & 1–16 \\
\texttt{GPQA-Main}    & Qwen3-8B-Thinking    & 1$\times$H100 & $n$-gram, EAGLE-3       & 32K & 1–16 \\
\texttt{AIME(22--24)} & GLM-4.5-Air    & 4$\times$H100 & $n$-gram, MTP (v0.11.1rc1)       & 32K & 1–4 \\
\texttt{GPQA-Main}    & GLM-4.5-Air     & 4$\times$H100 & $n$-gram, MTP (v0.11.1rc1)      & 32K & 1–4 \\
\bottomrule
\end{tabular}
\end{table*}

\begin{table*}[t]
\centering
\caption{
Model configurations for EAGLE and Draft-Model-based SD.
Each configuration specifies the target model, the associated draft (or auxiliary) model from Hugging Face,
and the number of speculative tokens proposed per decoding step.}
\label{appendix:model-config}
\begin{tabular}{l l c l}
\toprule
\textbf{Variant} & \textbf{Target Model} & \textbf{\# Draft Tokens} & \textbf{Draft / Auxiliary Model (Hugging Face)} \\
\midrule
\textbf{EAGLE} 
& Llama3.1-8B-Instruct & 3 & \texttt{yuhuili/EAGLE-LLaMA3.1-Instruct-8B} \\
& Llama3-70B-Instruct  & 3 & \texttt{yuhuili/EAGLE-LLaMA3-Instruct-70B} \\
\midrule
\textbf{EAGLE-3} 
& Llama3.1-8B-Instruct & 3 & \texttt{yuhuili/EAGLE3-LLaMA3.1-Instruct-8B} \\
& Qwen3-8B             & 3 & \texttt{AngelSlim/Qwen3-8B\_eagle3} \\
\midrule
\textbf{Draft-Model-Based} 
& Llama3-70B-Instruct  & 3 & \texttt{meta-llama/Llama-3.2-1B-Instruct} \\
& Qwen3-8B             & 3 & \texttt{Qwen/Qwen3-0.6B} \\
\bottomrule
\end{tabular}
\end{table*}
\pagebreak

\pagebreak
\section{Artifact Appendix}

\subsection{Abstract}

This artifact contains two components.
The first is a vLLM-based profiling suite that includes implementation and benchmarking scripts for speculative decoding in vLLM, covering end-to-end throughput, time breakdown, and acceptance rate experiments across multiple models and methods (N-gram, EAGLE, EAGLE-3, Draft Model, MTP). Full reproduction of this suite requires Linux, Python 3.10, CUDA 12.8, conda/uv, HuggingFace access to gated Llama-3 models, a one-time ShareGPT download, and NVIDIA H100-80GB GPUs (1 GPU for 8B models and 4 GPUs for 70B/106B-scale models).
The second component is a lightweight simulator that runs on any laptop with Python 3.10 and uses included pre-profiled acceptance traces to reproduce the paper’s upper-bound and combined-proposer analyses.
Evaluators can validate the artifact by running the profiling scripts to regenerate the main end-to-end results for Llama3.1-8B, or/and by running the simulator to reproduce the oracle upper-bound analyses. Expected outputs are PDF figures whose trends closely match Figure 1(a)-(d) , Figure 9, and Figure 11.

\subsection{Artifact check-list (meta-information)}

{\small
\begin{itemize}
  \item {\bf Algorithm: } N-gram, EAGLE, EAGLE-3, Draft Model, MTP speculative decoding; simulation of speculative decoding performance.
  \item {\bf Program: } (1) Python~3.10, vLLM (built from source per branch). (2) Python~3.10, \texttt{tqdm}, \texttt{pandas}, \texttt{matplotlib}.
  \item {\bf Compilation: } (1) \texttt{uv pip install -e .} via \texttt{rebuild\_env.sh}; CUDA~12.8. (2) \texttt{pip install -r requirements.txt}; no GPU required.
  \item {\bf Data set: } (1) InstructCoder, CNN/DailyMail, GSM8K, AIME, GPQA (HuggingFace, auto-downloaded); ShareGPT (one-time download, see \S\ref{sec:ae-install}). (2) Pre-profiled acceptance data included in \texttt{simulator/data/}.
  \item {\bf Run-time environment: } (1) Linux, CUDA~12.8, conda. (2) Any OS with Python~3.10.
  \item {\bf Hardware: } (1) 1x-4x H100-80GB GPU (2) No GPU required.
  \item {\bf Run-time state: } (1) No other process should occupy the target GPU(s). (2) N/A.
  \item {\bf Execution: } (1) \texttt{run-l3-8b.sh} for Llama3.1-8B end-to-end performance profiling. (2) \texttt{./simulate\_and\_plot.sh}.
  \item {\bf Metrics: } (1) Throughput speedup, acceptance rate, time breakdown x; (2) simulated speedup vs.\ batch size.
  \item {\bf Output: } (1) \texttt{scripts/results/run\_<timestamp>/} with \texttt{.jsonl} results and logs. (2) \texttt{results/\{proposer\}\_\{dataset\}\_speedup.csv} and PDF figures in \texttt{figures/llama3.1-8B/}.
  \item {\bf Experiments: } (1). \texttt{run-l3-8b.sh} will take ~24-36 hours to complete on 1x H100. (2). \texttt{simulate\_and\_plot.sh} will take ~30-40 minutes to finish.
  \item {\bf How much disk space required (approximately)?: } recommended to be in 500GB of free disk space.
  \item {\bf How much time is needed to prepare workflow (approximately)?: } (1) $\approx$30-40\,min for each environment building. (2) $<$5\,min.
  \item {\bf How much time is needed to complete experiments (approximately)?: } See Experiments above.
  \item {\bf Publicly available?: } Yes.
  \item {\bf Data licenses (if publicly available)?: } See HuggingFace model/dataset cards if applicable.
\end{itemize}
}

\subsection{Description}
\label{sec:ae-desc}

\subsubsection{How to access}

(1) vLLM suite: \href{https://github.com/SpecDecode-Bench/vllm}{https://github.com/SpecDecode-Bench/vllm}; see \texttt{README.md} for more information.
(2) Simulator: \href{https://github.com/SpecDecode-Bench/simulator}{https://github.com/SpecDecode-Bench/simulator}; see \texttt{README.md} for more information.

\subsubsection{Hardware dependencies}

(1) Profiling: 1x H100-80GB GPU for 8B models; 4x H100-80GB GPUs for larger models.
(2) Simulation: No GPU required; any machine with Python~3.10 suffices. 

\subsubsection{Software dependencies}

(1) \texttt{conda}, Python~3.10, \texttt{uv} (managed by
\texttt{rebuild\_env.sh}), \texttt{matplotlib}; HuggingFace access token for Llama-3 gated models.
(2) Python~3.10, \texttt{tqdm}, \texttt{pandas}, \texttt{matplotlib}
(installed via \texttt{pip install -r requirements.txt}).

\subsubsection{Datasets}

(1) All datasets except ShareGPT are downloaded automatically via
HuggingFace \texttt{datasets}. ShareGPT requires a one-time download
(see \S\ref{sec:ae-install}).
(2) Pre-profiled JSONL acceptance data for Llama-3.1-8B-Instruct on
H100 is included in \texttt{simulator/data/llama3.1-8B/} (gsm8k,
instructcoder, sharegpt, cnn).

\subsection{Installation}
\label{sec:ae-install}

\textbf{(1) vLLM suite: ShareGPT} (one-time):
\begin{verbatim}
huggingface-cli download  \
anon8231489123/ShareGPT_Vicuna_unfiltered \
    ShareGPT_V3_unfiltered_cleaned_split.json \
    --repo-type dataset --local-dir /path/to/data/
export SHAREGPT_PATH=/path/to/dataset
\end{verbatim}

\textbf{(1) vLLM suite: environment} (repeat per branch):
\begin{verbatim}
git checkout <branch>
ENV_DIR=/path/to/envs bash scripts/rebuild_env.sh
conda activate /path/to/envs
\end{verbatim}

\textbf{(2) Simulator}:
\begin{verbatim}
cd simulator/
conda create -n specbench_simulator python=3.10 -y
conda activate specbench_simulator
pip install -r requirements.txt
\end{verbatim}

\subsection{Experiment workflow}

\textbf{(1) vLLM profiling.}
Set \texttt{CUDA\_VISIBLE\_DEVICES} at the top of the target script,
then from \texttt{scripts/}:
\begin{verbatim}
export SHAREGPT_PATH=/path/to/sharegpt.json
bash run-l3-8b.sh    # 1 GPU
\end{verbatim}
Each script runs warmup, all datasets $\times$ methods, and calls
\texttt{vis\_speedup.py} to produce PDF figures in
\texttt{results/run\_<timestamp>/figures/}.

\textbf{(2) Simulator.}
\begin{verbatim}
cd simulator/
./simulate_and_plot.sh
\end{verbatim}
This runs EAGLE-3, N-gram, and combined proposers across all four
datasets and saves CSV results to \texttt{results/} and PDF figures to
\texttt{figures/llama3.1-8B/}.

\subsection{Evaluation and expected result}

\textbf{(1) vLLM profiling.}
After running \texttt{run-l3-8b.sh}, we expect the same end-to-end performance results to be reproduced for Figure~1\,(a)--(d).

\textbf{(2) Simulator.} We expect the same results to be reproduced for Figure 9 and Figure 11.

\subsection{Experiment customization}

\textbf{(1) vLLM profiling.}
Set \texttt{num\_reqs=5} and \texttt{batch\_sizes="1 16"} for a
$<$10\,min smoke test (use a non-reasoning script).
Individual datasets and methods can be toggled by editing the
\texttt{for} loops.

\textbf{(2) Simulator.}
Individual proposers, datasets, prediction methods, and batch sizes
can be configured via CLI flags to \texttt{main.py}. Refer to the README on GitHub for more information.

\subsection{Notes}

For profiling, each script run writes to a fresh timestamped directory. Each branch requires its own virtual environment.

\end{document}